\documentclass[lettersize,journal]{IEEEtran}
\usepackage{amsmath,amsfonts}
\usepackage{algorithmic}
\usepackage{algorithm}
\usepackage{array}
\usepackage[caption=false,font=normalsize,labelfont=sf,textfont=sf]{subfig}
\usepackage{textcomp}
\usepackage{stfloats}
\usepackage{url}
\usepackage{verbatim}
\usepackage{graphicx}
\usepackage{cite}
\usepackage[normalem]{ulem}
\usepackage{tabularray}
\usepackage{CJKutf8}
\usepackage{color}
\usepackage{cite}
\usepackage{diagbox} 
\usepackage{hyperref}
\usepackage{pifont}

\begin{document}
\begin{CJK}{UTF8}{gbsn}

\title{VTire: A Bimodal Visuotactile Tire with High-Resolution Sensing Capability }
\author{Shoujie Li,~\IEEEmembership{Student Member,~IEEE},
         Jianle Xu, Tong Wu, Yang Yang, Yanbo Chen, Xueqian Wang,~\IEEEmembership{Member,~IEEE}, Wenbo Ding,~\IEEEmembership{Member,~IEEE},
         Xiao-Ping Zhang,~\IEEEmembership{Fellow,~IEEE}
\thanks{
This work was supported by National Key R\&D Program of China grant (2024YFB3816000),
Shenzhen Key Laboratory of Ubiquitous Data Enabling (No. ZDSYS20220527171406015), Shenzhen Science and Technology Program (JCYJ20220530143013030), Guangdong Innovative and Entrepreneurial Research Team Program (2021ZT09L197), National Natural Science Foundation of China (62104125, 62003188), Tsinghua Shenzhen International Graduate School-Shenzhen Pengrui Young Faculty Program of Shenzhen Pengrui Foundation (No. SZPR2023005) and Meituan. 
\textit{Shoujie Li and Jianle Xu contributed equally to this
work.} (Corresponding author: Xueqian Wang  \& Wenbo Ding, wang.xq@sz.tsinghua.edu.cn \& ding.wenbo@sz.tsinghua.edu.cn)}

\thanks{Shoujie Li, Jianle Xu, Tong Wu,  Yanbo Chen, Xueqian Wang, Wenbo Ding and Xiao-Ping Zhang are with Shenzhen International Graduate School, Tsinghua University, Shenzhen 518055, China.}
\thanks{Wenbo Ding is also with the RISC-V International Open Source Laboratory, Shenzhen 518055, China.}
\thanks{Yang Yang is with the Department of Mechanics Science and Engineering, Sichuan University, Chengdu 610065, China }}

\markboth{Journal of \LaTeX\ Class Files,~Vol.~14, No.~8, August~2021}%
{Shell \MakeLowercase{\textit{et al.}}: A Sample Article Using IEEEtran.cls for IEEE Journals}



\maketitle
\begin{abstract}
Developing smart tires with high sensing capability is significant for improving the moving stability and environmental adaptability of wheeled robots and vehicles. However, due to the classical manufacturing design, it is always challenging for tires to infer external information precisely. To this end, this paper introduces a bimodal sensing tire, which can simultaneously capture tactile and visual data. By leveraging the emerging visuotactile techniques, the proposed smart tire can realize various functions, including terrain recognition, ground crack detection, load sensing, and tire damage detection.  Besides, we optimize the material and structure of the tire to ensure its outstanding elasticity, toughness, hardness, and transparency. In terms of algorithms, a transformer-based multimodal classification algorithm, a load detection method based on finite element analysis, and a contact segmentation algorithm have been developed. Furthermore, we construct an intelligent mobile platform to validate the system's effectiveness and develop visual and tactile datasets in complex terrains. The experimental results show that our multimodal terrain sensing algorithm can achieve a classification accuracy of 99.2\%, a tire damage detection accuracy of 97\%, a 98\% success rate in object search, and the ability to withstand tire loading weights exceeding 35 kg. In addition, we open-source our algorithms, hardware, and datasets at \href{https://sites.google.com/view/vtire}{https://sites.google.com/view/vtire}.

\end{abstract}


\section{Introduction}

\IEEEPARstart{T}{ire} 
 is a crucial actuating component, which has a wide range of applications in vehicles~\cite{10646374} and wheeled bipedal robots~\cite{10597662}. As the only part of the robot in contact with the ground, tires could enhance the driving stability as well as serve as the most dependable source of information regarding the ground. However, due to the classical manufacturing and sensing technology, the existing tires only obtain limited information such as load, speed, acceleration, etc.~\cite{lee2017intelligent}, which cannot realize pixel-level texture sensing. Designing a smart tire with high-resolution tactile sensing ability can solve the problem of terrain sensing under complex scenarios and realize more functions, including ground crack detection, ground object search, and tire damage detection, dramatically improving the robot's or vehicle's sensing ability.

\begin{figure}
	\centering
  \includegraphics[width=0.45\textwidth]{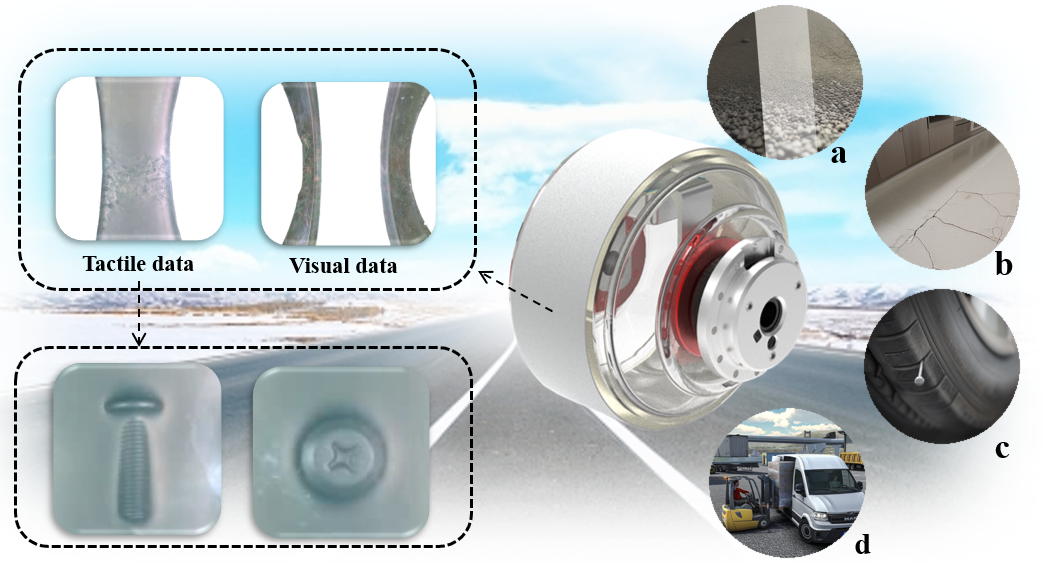}
	\caption{Introduction to the bimodal Tire. Functions that can be achieved with tires: (a) Terrain classification in complex scenes; (b) Ground cracks and objects searching; (c) Tire damage detection; (d) Weight-loaded detection. } \label{fig:1}
\end{figure}

To enhance the sensing ability of tires, researchers usually adopt a variety of force~\cite{xu2020tire,maurya20203d} and optical sensors~\cite{tuononen2011laser,tuononen2009board,matsuzaki2010optical} inside the tire, which could provide the tire with the ability to sense its states, such as load, tension, and so on. Nevertheless, due to the obstruction of the tire casing, such methods can not obtain external information, such as the road texture, cracks, etc. With advancements in tactile perception and optical imaging technologies, a novel technique called visuotactile perception ~\cite{abad2020visuotactile,lu2024dexitac,10358360} has emerged. This technology detects tactile information by observing surface deformations through a camera, offering high resolution and extensive sensing coverage. However, tires' opacity, hardness, and toughness pose difficulties in integrating visuotactile perception technology.

In this paper, by improving tires' materials, mechanical structure, and manufacturing process, we propose a bimodal smart tire named VTire based on visuotactile sensing techniques. As shown in Fig.~\ref{fig:1}, VTire overcomes the bottlenecks in the resolution of tactile sensing and sensing area of traditional smart tires and realizes the functions of sensing ground texture, cracks, and bumps, which are challenging for traditional smart tires. The contributions of this paper are as follows:

\begin{itemize}

\item {\bf Innovative manufacturing process:} We propose a tire manufacturing method that capitalizes on visuotactile perception, ensuring high-resolution tactile data acquisition. Based on the unique structural design, our tires acquire both tactile and visual information, which enhances the perception of the external environment.

\item {\bf Advanced sensing algorithms:} Our research introduces a series of sophisticated sensing algorithms tailored for smart tires. These algorithms include a transformer-based multimodal classification algorithm, a load detection algorithm using finite element analysis, and a contact segmentation algorithm.

\item {\bf Comprehensive dataset:} To evaluate the performance of our algorithms, we assembled a diverse dataset. This dataset encompasses various terrains, surface textures, tire damage scenarios, and ground cracks, providing a comprehensive evaluation framework.

\item {\bf Numerous validation experiments:} To verify the performance of the tires, we not only built a smart mobile platform but also designed several real-world experiments. Results indicate remarkable accuracy rates, including 0.75 kg weight sensing accuracy, 98\% crack segmentation accuracy, 98\% object detection success rate, and 97\% damage detection accuracy. These results underscore the practical viability of our approach.

\end{itemize}

\section{Related Work}

The design of smart tires can be divided into contact sensing and non-contact sensing. contact sensing mainly includes accelerometers~\cite{singh2015estimation,barbosa2021lateral},  surface acoustic wave (SAW) sensors~\cite{oh2012development,zhang2004design}, piezoelectric sensors~\cite{4453936,10552090}, strain sensors~\cite{mendoza2020strain,yunta2019influence}, and fiber bragg grating (FBG) strain sensors~\cite{gubaidullin2019microwave}.  The contact sensors are mounted on the tire's inner wall and can accurately obtain the pressure and tension when the tire is in contact with the ground.  However, the spatial resolution of these sensors is low, and since the sensors are pasted on the inner side of the tire, they are easy to detach and malfunction after long-distance movement. Non-contact sensors can acquire tire deformation without touching the tire's surface, which greatly avoids the risk of the sensor falling off and improves the sensor's service life. Typical non-contact sensors include optical sensors~\cite{tuononen2011laser,tuononen2009board,matsuzaki2010optical} and ultrasonic sensors~\cite{longoria2019wheel}. Most optical sensors use single-point laser sensors or detect the offset of a feature point to obtain contact information. Due to the obstruction of the tire casing, it is difficult for these methods to obtain the texture, deformation, and other information of the contact road surface, which greatly compresses the application range of smart tires, as shown in Table. \ref{tire}. To solve this problem, Hu \textit{et al.}~\cite{10354894} proposed a visual tire that uses transparent acrylic as the casing. While this solution can achieve terrain detection, it uses a visual solution susceptible to the external environment.  Furthermore, the acrylic casing not only makes the vehicle prone to skidding but also makes the tire lose its elasticity. When the vehicle passes through some rough, dirty road surface, it is easy to wear on the shell, thus affecting the detection effect.

\begin{table}[]
{
\caption{Comparison of current smart tire functions}
\begin{center}
\begin{tabular}{m{1.2cm}<{\centering}|p{1cm}<{\centering}|m{1.5cm}<{\centering}|m{0.8cm}<{\centering}|p{0.8cm}<{\centering}|p{0.8cm}<{\centering}}

\hline
Ref & Pressure & Deformation & Crack & Terrain & Damage \\ \hline

Ref.\cite{maurya20203d}    &   \ding{52}       &      \ding{56}         &    \ding{56}     &      \ding{56}     &     \ding{56}     \\ 
Ref.\cite{matsuzaki2010optical}    &    \ding{52}      &       \ding{52}      &     \ding{56}    &     \ding{56}      &    \ding{56}      \\

Ref.\cite{longoria2019wheel}   &     \ding{56}       &     \ding{52}        &      \ding{56}   &      \ding{56}     &   \ding{56}       \\
Ref.\cite{kim2020road}    &   \ding{52}       &     \ding{56}          &    \ding{56}     &   \ding{52}      &     \ding{56}     \\
Ref.\cite{khaleghian2017terrain}   &      \ding{56}      &      \ding{56}       &    \ding{56}     &   \ding{52}      &    \ding{56}      \\

Ref.\cite{eun2016highly}    &      \ding{56}      &   \ding{52}          &    \ding{56}     &     \ding{56}      &    \ding{56}      \\

 \bf {Ours}   &   \ding{52}       &     \ding{52}        &    \ding{52}    &     \ding{52}     &     \ding{52}     \\ \hline

\end{tabular}
\end{center}
\label{tire}
}
\end{table}

\section{Hardware Design}

\begin{figure*}
	\centering
  \includegraphics[width=0.9\textwidth]{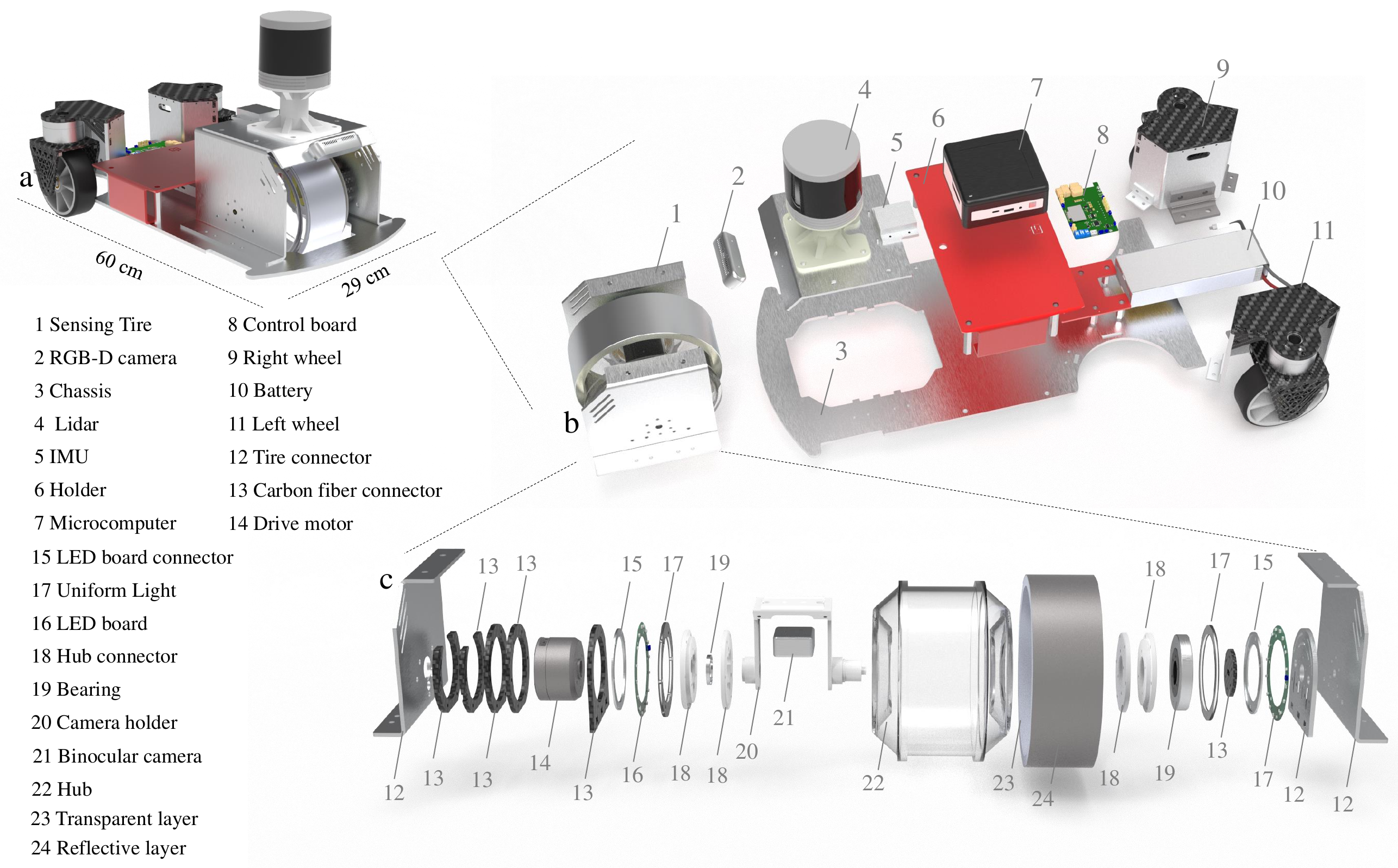}
	\caption{Hardware structure diagram. (a) Side view of the motion platform; (b) The mobile platform exploded view; (c) VTire exploded view. } \label{fig:2}
\end{figure*}

To improve the sensing ability and mechanical strength, we optimize the tires in terms of structure and material. Besides, to demonstrate the performance of the smart tires, we build an intelligent mobile platform, as shown in Fig.~\ref{fig:2}.

\subsection{System Structure}

\begin{figure}
	\centering
  \includegraphics[width=0.45\textwidth]{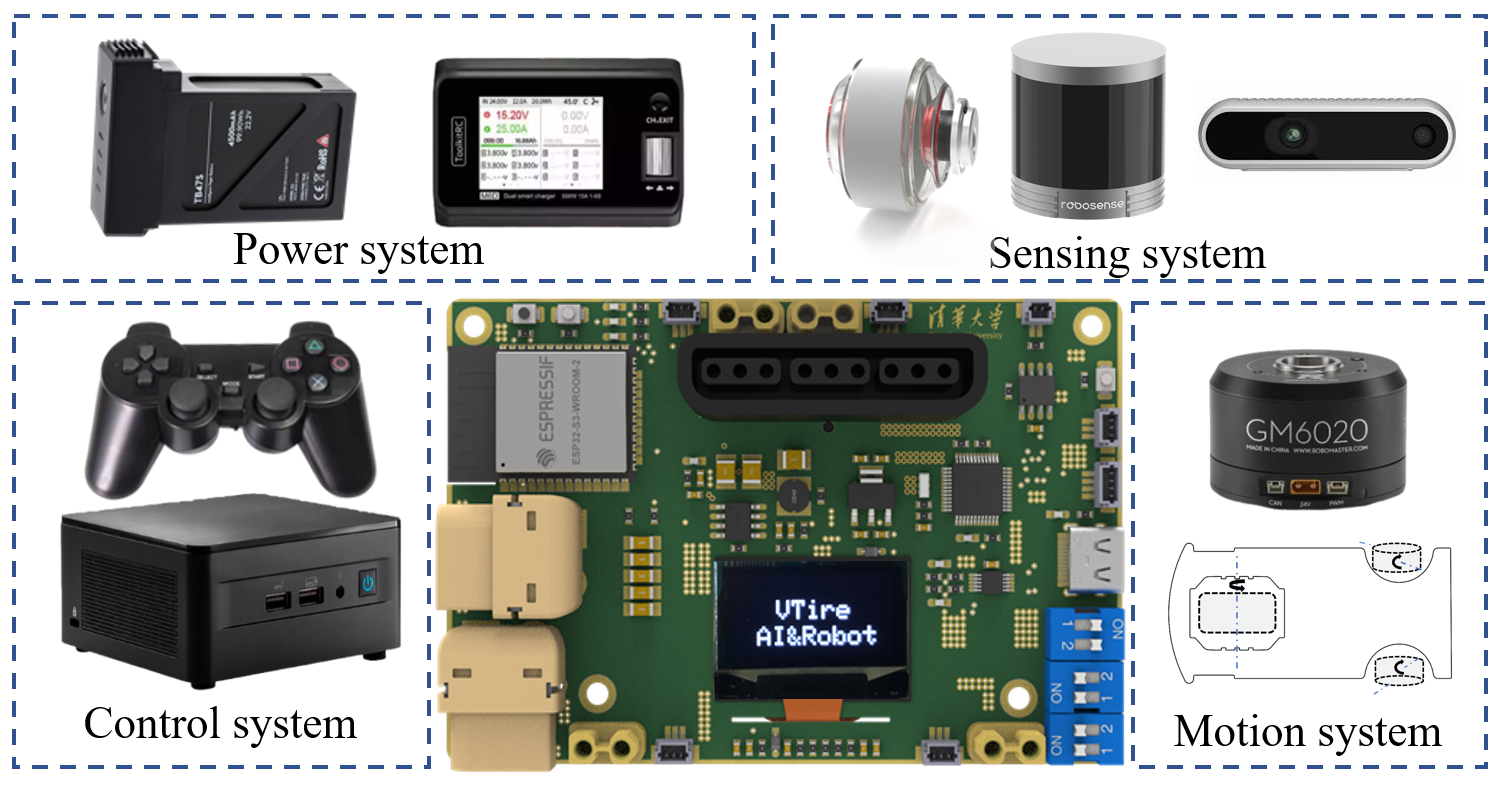}
	\caption{The mobile platform framework diagram. The mobile platform consists of a power system, a sensing system, a control system, and a motion system.}  \label{fig:ha}
\end{figure}

The mobile platform is 60 mm long and 29 mm wide and consists of a power system, a sensing system, a control system, and a motion system.  To ensure structural stability, the body is made of metal and carbon fiber with a high weighing capacity that can carry adults over 70 kg.  The battery is DJI TB47, and the voltage regulator module is ToolkitRC, which can output up to 40A current. The sensing system consists of Vtire, lidar (OS0-128), and RGB-D camera (Realsense D435i), which enables real-time mapping and terrain classification in different environments. The control system comprises a remote control handle and a Next Unit of Computing NUC. All systems are connected through our designed main control board, as shown in Fig.~\ref{fig:ha}. Vtire not only performs the sensing function but is also used to provide the driving force. Because the front wheels have larger friction, this front-wheel drive improves the platform's ability to cross obstacles. The rear wheel steering configuration makes the platform more agile with a smaller turning radius.

\subsection{Proposed Smart Tire }

Like the classic visuotactile sensors, smart tires are composed of a sensing skin, a lighting system, and a vision system. The difference, however, is that the design of smart tires needs to consider their transparency, elasticity, load capacity, and other metrics, making them highly challenging in terms of dimensions, fabrication process, and material selection.

\subsubsection{Sensing skin }

\begin{figure*}
	\centering
  \includegraphics[width=0.9\textwidth]{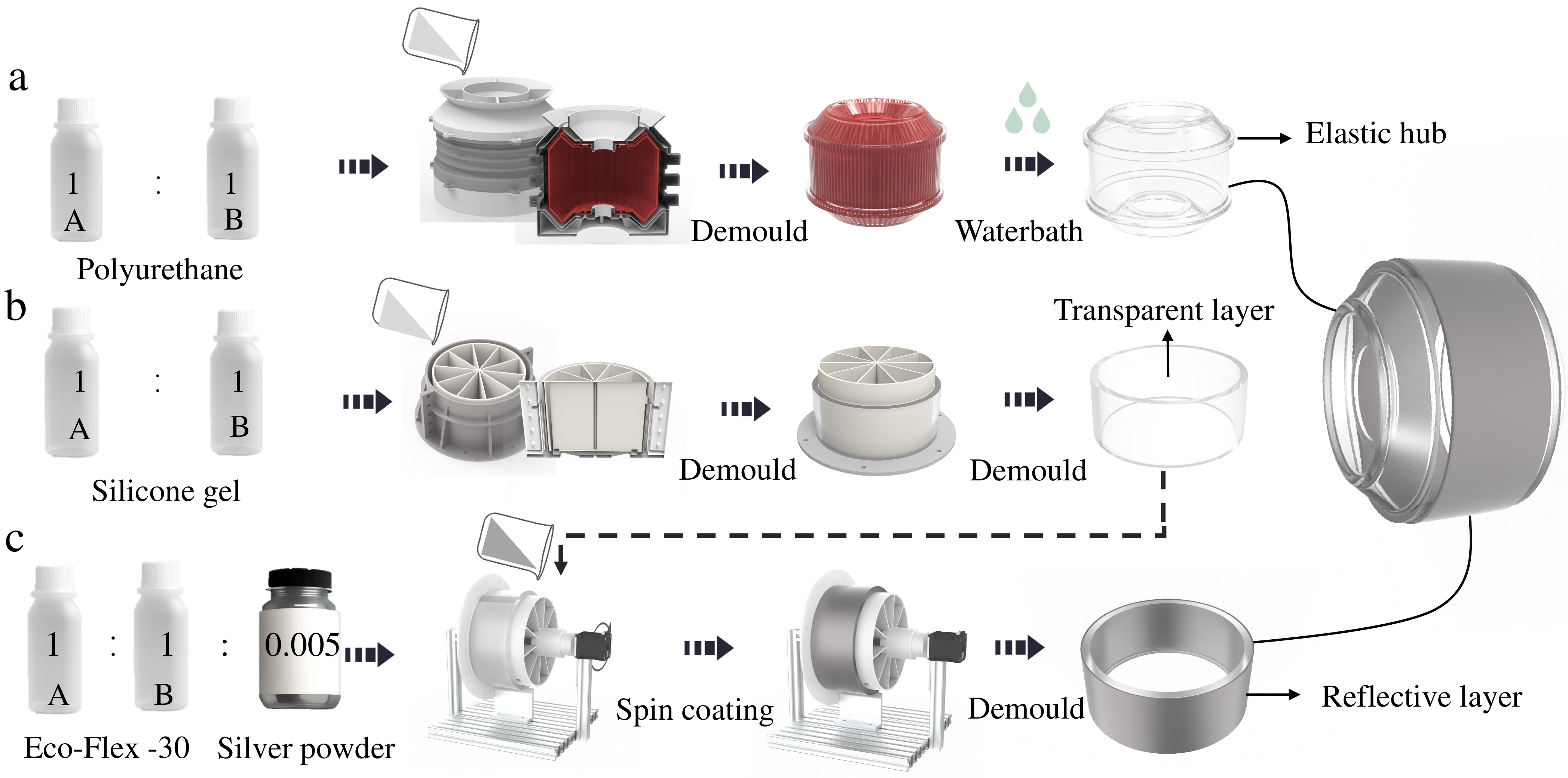}
	\caption{Fabrication processes of (a) the wheel hub; (b) the transparent layer; (c) the reflective layer. } \label{fig:3}
\end{figure*}

The sensing skin is the heart of the tire, which consists of a transparent elastic hub, a transparent layer, and a reflective layer. Transparent elastic wheels not only support the robot body but also have the function of vibration damping. So, transparent wheels need to be elastic, tough, and transparent. After considering silicone, rubber, acrylic, glass, and other materials, we use a polyurethane (PU) material, which has a lower price, higher hardness, and transparency and is widely used in skateboard wheels, omnidirectional wheels, and other fields.
The tire adopts non-sealed structures and the manufacturing process of the hub is illustrated in Fig.~\ref{fig:3}a. The molds are all made by 3D printing. Polyurethane materials have a high viscosity during the demolding process. To ensure that the surface of the mold is smooth, a special release agent is used, and clear tape is applied to key areas. In addition, we used a water-soluble material (Polyvinyl alcohol (PVA)) as an internal support (red part in Fig.~\ref{fig:3}a), which softens when dissolved in water so that the hub can be unmolded smoothly.

To ensure the perception ability of the tire, we also design the elastic transparent layer (Fig.~\ref{fig:3}b) and a reflective layer (Fig.~\ref{fig:3}c). For the elastic transparent layer, we use low-hardness elastic silicone and design a removable mold to realize the transparent layer. We use Eco-Flex 30 as the material for the reflective layer and mix it with silver powder to enhance the texture perception. In addition, to ensure the homogeneity of the outermost layer structure, we design a squeegee to control the thickness of the outermost film of the tire through the squeegee and to make the outermost layer of the tire more homogeneous through rotation. We open-source the production process and hardware, detailed on the website \href{https://sites.google.com/view/vtire}{https://sites.google.com/view/vtire}.

\subsubsection{Imaging system }

\begin{figure}
	\centering
  \includegraphics[width=0.45\textwidth]{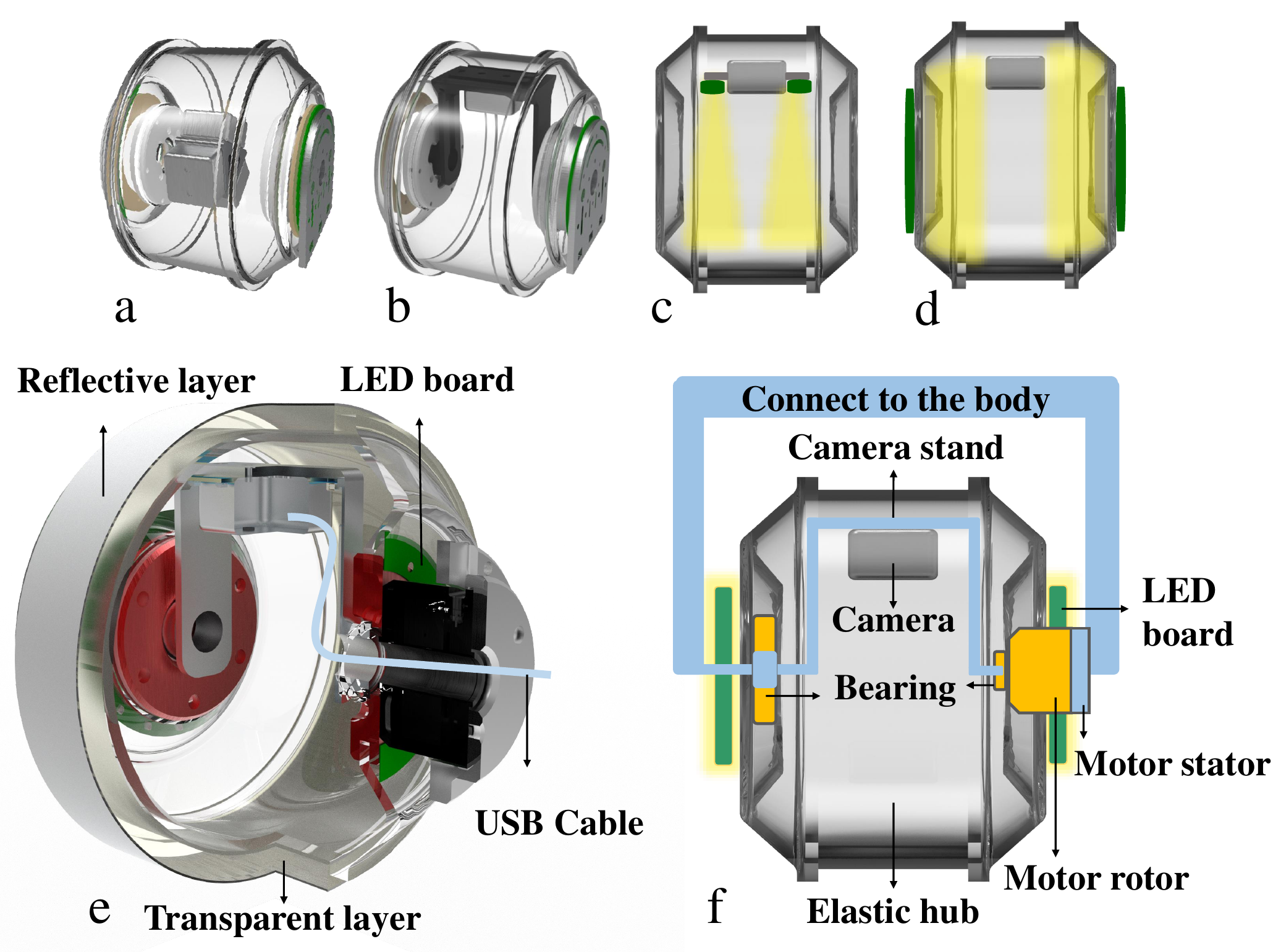}
	\caption{ Comparison of different solutions. (a) Multi-camera solution; (b) Single camera solution; (c) Top lighting solution; (d) Side lighting solution.  (e) Section view. (f) Connection of components.} \label{fig:4}
\end{figure}

For the vision system design, we consider two options: a multi-camera solution (Fig.~\ref{fig:4}a) and a single-camera structure (Fig.~\ref{fig:4}b). The multi-camera one can monitor the status of the whole tire in real-time, but the design structural complexity is high, the main issues are as follows:

\begin{itemize}

\item {\bf Installation difficulties:} Due to the limited space inside the tire, factors such as the camera's focal length and imaging range must be considered during installation. Since we are using binocular imaging technology, the camera has a minimum detection distance of 12cm, which is almost the limit for the tire. Moreover, the presence of the tire's central axis will further compress the available space for camera installation when multiple cameras are deployed.

\item {\bf Data transmission difficulties:} In the single-camera solution, the camera does not rotate with the tire, so the camera's data can be stably transmitted to the computer. However, the cameras rotate with the tire when using the multi-camera solution, making transmitting data to the computer difficult. On the one hand, the data cables may become entangled due to the tire's rotation, and on the other hand, rotation can affect the stability of the camera's data transmission.

\item {\bf Poor reliability:} With the multi-camera solution, the cameras rotate with the tire, and the solid centrifugal force can easily damage the cameras.
\end{itemize}

For the above reasons, we abandoned the multi-camera solution. The single-camera solution uses a bearing + bracket structure to fix the camera on the body and prevent the camera from rotating with the tire.
To obtain the depth information inside the tire, we adopt the binocular depth imaging technology, using Intel's Realsense D405 as the image acquisition device and obtaining the texture and deformation information of the tire in real time through the binocular depth reconstruction algorithm. In addition, the camera has a high image rate, which utilizes a global shutter and can reach a frame rate of 90 fps.

To solve the power supply and data transmission problem, we used hollow motors as VTire drivers so the USB cable could pass through the hollow shaft without getting tangled, as shown in Fig.~\ref{fig:4}e. The blue part of Fig.~\ref{fig:4}f represents the part where the VTire is connected to the body. It is fixed during the rotation of the V-tire, while the gray part will rotate driven by the motor rotor. Since the camera stand is fixed to the body, it always keeps the field of view facing the ground. 
In addition to the difficulty of deployment, there are a few reasons why we went with a single-camera solution:

\begin{itemize}

\item {\bf Cost issue:} The multi-camera solution would significantly increase our costs.

\item {\bf Structural issue: } When the tire moves, only a part of the area will be in contact with the ground so that the single camera solution can meet the tactile information collection needs when in contact with the ground and the multi-camera solution doesn't get more contact information.

\item {\bf Application scenario issue:} Our application scenarios are mainly aimed at low-speed scenarios, such as ground crack detection, ground small object search, and other tasks. We have chosen a camera with a frame rate of 90 fps (traditional cameras only have 30 fps), which can meet the needs of these tasks. If it is for high-speed scenarios, we can use cameras with higher frame rates, such as event cameras.
\end{itemize}

\subsubsection{Lighting system }

For the lighting system, we have also attempted several different schemes. To reduce the effect of reflections inside the camera, we choose the solution shown in Fig.~\ref{fig:4}d. We can achieve a uniform lighting effect by installing LED light rings on both sides of the tire.

\section{Algorithm Design}

\subsection{Multimodal Terrain Classification Algorithm}

\begin{figure}
	\centering
  \includegraphics[width=0.45\textwidth]{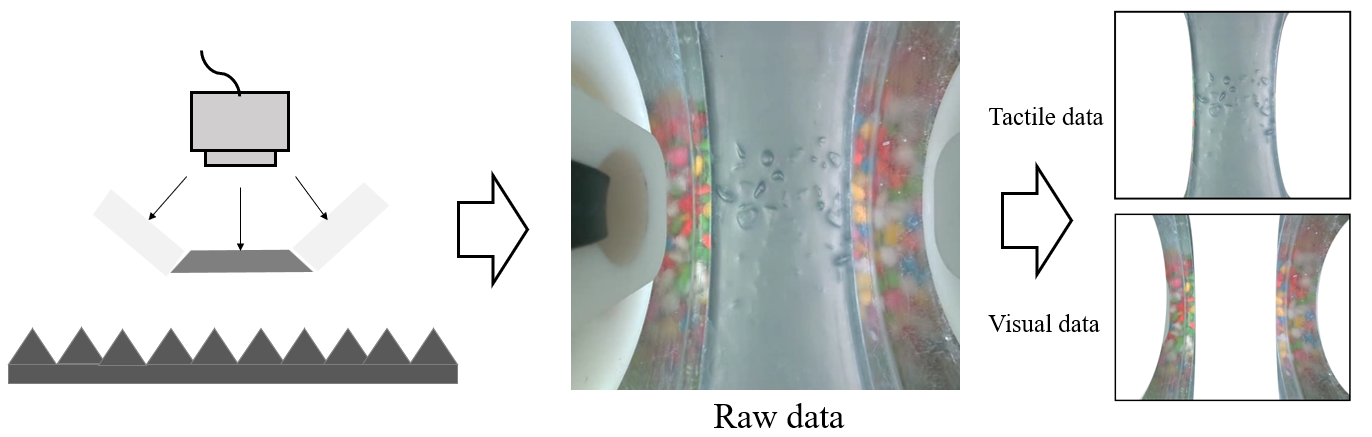}
	\caption{ Schematic of bimodal perception.  } \label{fig:bimodal}
\end{figure}

Vision is the most direct means for robots to perceive the external environment. However, factors such as smoke, brightness, and dust can significantly impact the visual approach to terrain classification~\cite{9882387}. In contrast, tactile perception provides greater stability in such scenarios.
Compared with classic visuotactile sensors, VTire can obtain tactile and visual data at the same time. As shown in Fig.~\ref{fig:bimodal}, the center area of the tires is covered with sensing skin, and the side area is transparent.
The transparent areas allow us to perceive external visual information while acquiring tactile information. In addition, cameras on the outside of the vehicle can also provide clearer visual information compared to the inside of the tire.

\begin{figure}
	\centering
  \includegraphics[width=0.45\textwidth]{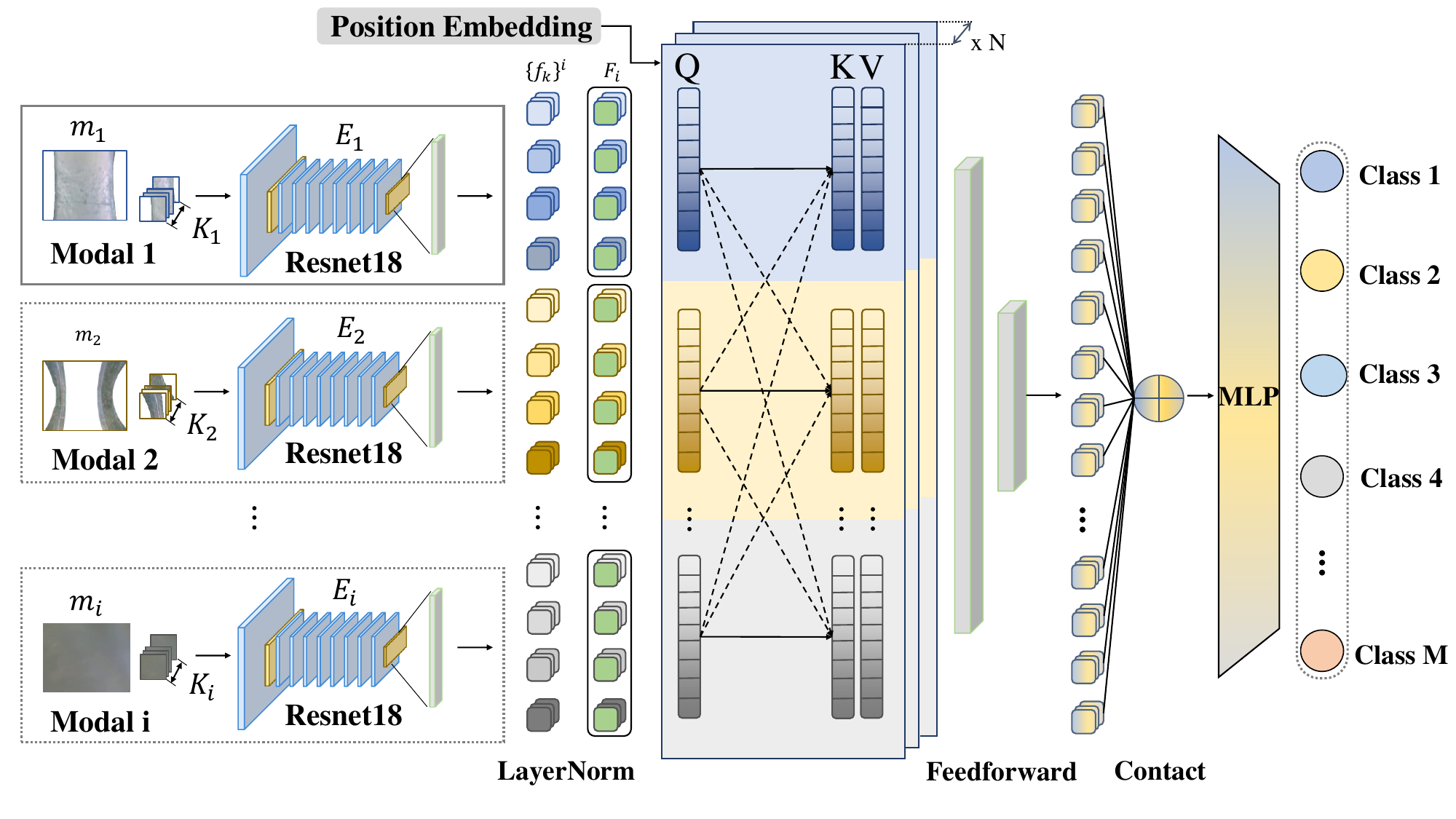}
	\caption{  Transformer-based multimodal terrain classification algorithm.  } \label{fig:network}
\end{figure}

To better integrate these data, we propose a multimodal classification algorithm, which adopts a transformer~\cite{vaswani2017attention} network as the body of the algorithm, as shown in Fig.~\ref{fig:network}. By adjusting the inputs to the network, it is possible to classify the data directly from a single modality to multiple modalities. Given a series of different modalities $m_i$, $i=0,1,...,M-1$,
which may include optical images of the terrain, tactile feedback from the surface, or partial observations such as dark or smoky images, we employ a set of modality-specific encoders $\{E_i\}$ to extract features from each corresponding modality. Instead of extracting a global feature for each modality, we partition each modality into $K_i$ fragments and use encoder $E_i$ to extract $K_i$ fragmented features. Since data distribution of these modalities may differ significantly, complicating modality fusion and training process, we apply LayerNorm~\cite{Ba2016LayerN} to each modality to alleviate this issue. The process of feature extraction can be formulated as follows:
\begin{equation}
    \{\boldsymbol{f}_k\}^i = \text{LayerNorm}(E_i(\text{Seg}(m_i))).
\end{equation}

To effectively utilize the information within each modality and across different modalities, we employ self-attention and cross-attention mechanisms~\cite{Chen2022VisuoTactileTF} to fuse the extracted features. Given the obtained features $\{\boldsymbol{f}_k\}^i$, we initially concatenate them along the fragment dimension, followed by concatenation along the modality dimension. Positional embedding is then added to the feature map. From the tokenized features $\boldsymbol{F}$ we can derive the following formulation:
\begin{equation}
    \boldsymbol{F} = \left [ \begin{matrix} \boldsymbol{F}_0 & \boldsymbol{F}_1 & \cdots & \boldsymbol{F}_{M-1}\end{matrix} \right ],
\end{equation}
where $\boldsymbol{F}_i$ denotes the concatenated feature of $i$th modality. To calculate self-attention and cross-attention, we pass $\boldsymbol{F}$ through a linear layer to derive the queries $\{\boldsymbol{Q}_j\}$, keys $\{\boldsymbol{K}_j\}$ and values $\{\boldsymbol{V}_j\}$ for each attention head $j=1,2,...,N$:
\begin{equation}
    \boldsymbol{Q}_j = \boldsymbol{W}_j^Q \boldsymbol{F} = \left [ \begin{matrix} \boldsymbol{Q}_{j,0} & \boldsymbol{Q}_{j,1} & \cdots & \boldsymbol{Q}_{j,M-1} \end{matrix} \right ],
\end{equation}
\begin{equation}
    \boldsymbol{K}_j = \boldsymbol{W}_j^K \boldsymbol{F} = \left [ \begin{matrix} \boldsymbol{K}_{j,0} & \boldsymbol{K}_{j,1} & \cdots & \boldsymbol{K}_{j,M-1} \end{matrix} \right ],
\end{equation}
\begin{equation}
    \boldsymbol{V}_j = \boldsymbol{W}_j^V \boldsymbol{F} = \left [ \begin{matrix} \boldsymbol{V}_{j,0} & \boldsymbol{V}_{j,1} & \cdots & \boldsymbol{V}_{j,M-1} \end{matrix} \right ],
\end{equation}
where $\boldsymbol{W}_j^Q$, $\boldsymbol{W}_j^K$, and $\boldsymbol{W}_j^V$ denote the weight matrices for the query, key, and value, respectively. The terms $\boldsymbol{Q}_{j,i}$, $\boldsymbol{K}_{j,i}$, $\boldsymbol{V}_{j,i}$ denote the query, key and value of head $j$ corresponding to the $i$th modality. The attention for head $j$ is then calculated using the following equation:
\begin{equation}
\begin{aligned}
    \boldsymbol{A}_j &=\!\! \boldsymbol{V}_j\!\! \times \!\!\text{softmax}((\boldsymbol{K}_j)^T \boldsymbol{Q}_j / \sqrt{d}) \\
        &=\!\! \boldsymbol{V}_j\!\! \times \!\!\text{softmax}(\begin{bmatrix}
 \underline{(\boldsymbol{K}_{j,0})^T \boldsymbol{Q}_{j,0}} & (\boldsymbol{K}_{j,0})^T \boldsymbol{Q}_{j,1} & \cdots\\
 (\boldsymbol{K}_{j,1})^T \boldsymbol{Q}_{j,0} & \underline{(\boldsymbol{K}_{j,1})^T \boldsymbol{Q}_{j,1}} & \cdots\\
 \vdots & \vdots & \ddots
\end{bmatrix} / \sqrt{d}),
\end{aligned}
\end{equation}
where $d$ is the dimension of the multimodal features. The self-attention and cross-attention mechanisms are as follows: the diagonal elements represent self-attention, where self-generated queries are used to draw self-generated keys. Conversely, the non-diagonal elements represent cross-attention, wherein queries and keys are generated from different modalities to extract features. Subsequently, we pass the attention outputs through a feedforward network and concatenate all tokens into a fused feature. In the end, a classification head is employed to produce the output.

In the practical implementation, we preprocess the raw images into 16$\times$16 patches and select ResNet18~\cite{He2015DeepRL} as the backbone for the encoder $E_i$. This choice is due to ResNet18's efficiency in extracting diverse features and its capacity for facilitating rapid training. We configure the feature dimension to be 384. To promote effective modality fusion, we utilize 4 attention heads, ensuring diverse attention patterns across modalities. We also find that a single attention block suffices for all tasks. For the classification task, we employ a straightforward MLP comprising a hidden layer followed by a softmax layer. During training, we minimize the cross-entropy loss function using the batch size of 32.

\subsection{Load Weight Sensing Algorithm}

\begin{figure}
	\centering
  \includegraphics[width=0.4\textwidth]{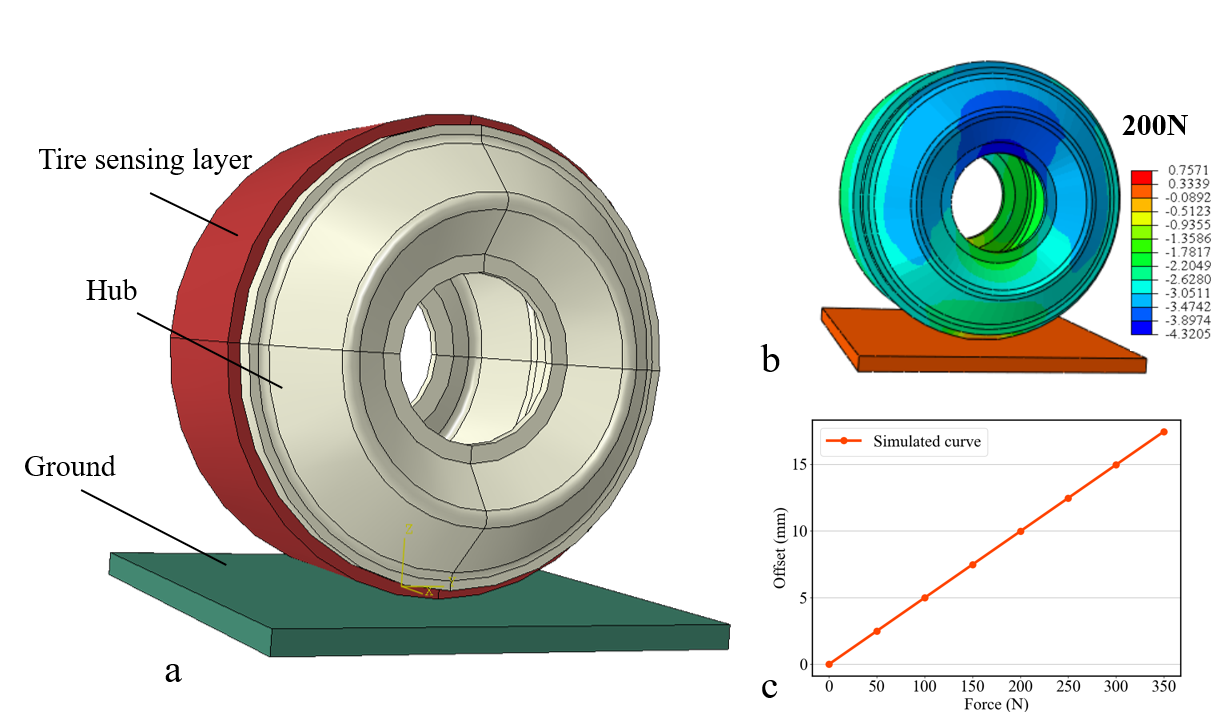}
	\caption{Simulation results. (a) Initial FEA simulation configuration; (b) Displacement deformation under 200 N; (c) Simulated relationship between deformation and force. } \label{fig:weight1}
\end{figure}

The load weight of a robot is directly correlated with the deformation experienced by its tires. To explore this relationship, we adopt the Abaqus/Standard Finite Element Analysis (FEA) simulation. This approach allowed us to model the tire accurately under various load conditions. The loading process's configuration and the simulation setup are depicted in Fig.~\ref{fig:weight1}a.

\begin{itemize}

\item {\bf Material properties:} 
Based on the experimental measurement results, the mechanical parameters of the hub and tire sensor layer are determined as $E=0.1973$ MPa, $\nu=0.48$, and $E=24.06$ MPa, $\nu=0.49$, respectively. Notably, the ground is considered a rigid body in the simulation.

\item {\bf Interaction and loading:} 
Interactions between the hub and sensor layer and between the hub and ground are defined as constraints within the model. The applied load is centrally located on the hub and directed towards the ground.

\item {\bf Mesh:} 
 For the mesh, we employ a simplified integration hexahedral 8-node element (C3D8R) across the model's three components.

 \item {\bf Simulated results:} 
We integrate Python scripting with Abaqus to dynamically adjust the applied force, enabling us to derive the deformation-force relationship. 
\end{itemize}
Here, the deformation field at an applied force of 200N is depicted in Fig.~\ref{fig:weight1}b, while Fig.~\ref{fig:weight1}c illustrates the corresponding deformation-force curve.

\subsection{Floor Crack and Contact Object Segmentation Algorithm}

\begin{figure}
	\centering
  \includegraphics[width=0.45\textwidth]{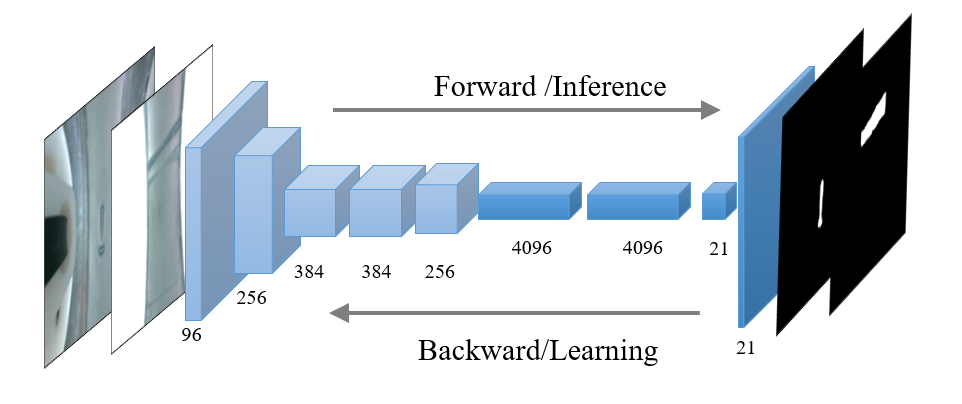}
	\caption{Floor crack and contact object segmentation algorithm.  } \label{fig:segment}
\end{figure}

Compared with vision, touch is more stable and robust. It is difficult to detect some small or transparent objects using only vision. Besides, crack detection could be challenging with some richly textured floors. To solve this problem, we adopt the FCN~\cite{long2015fully} to segment the areas of VTire in contact with objects, of which the network structure is shown in Fig.~\ref{fig:segment}.

\section{Experiments}

\subsection{Bimodal Terrain Classification (VTire Bimodal Data)}

\begin{figure*}[t]
	\centering
  \includegraphics[width=0.95\textwidth]{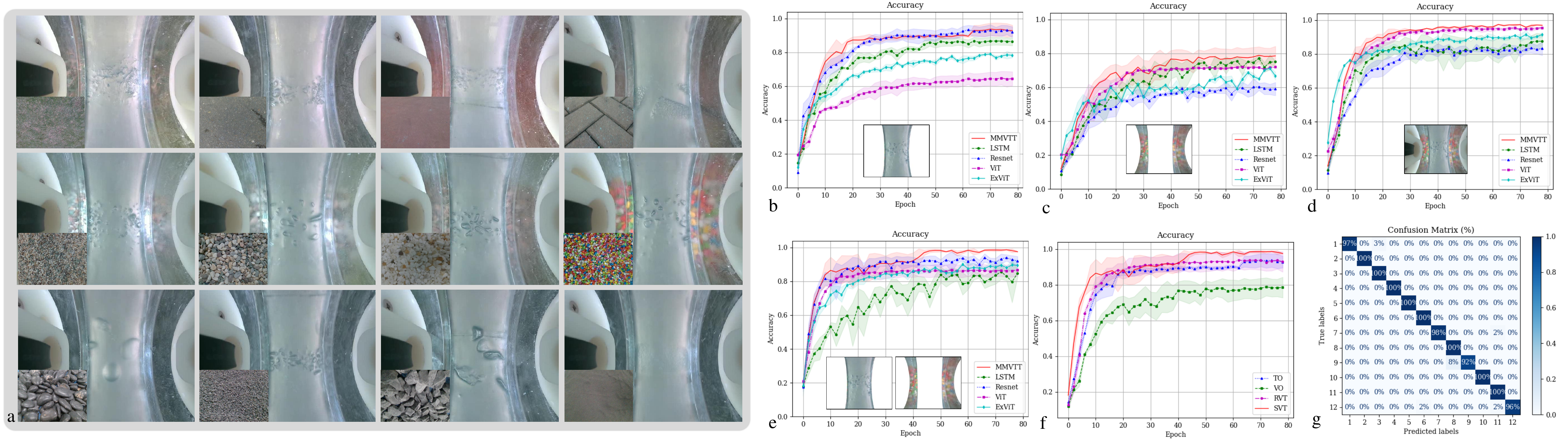}
	\caption{ Bimodal terrain classification. (a) Raw visuotactile data in different terrains; (b) The classification result of tactile data only (TO);  (c) The classification result of the sensor's visual data only (VO); (d)  The classification result of the raw visuotactile data (RVT); (e) The classification result of segmented visuotactile data (SVT); (f) The classification result of our proposed network in different modalities; (g) The confusion matrix of our proposed network for segmented visual image data. } \label{fig:Bimodalexp}
\end{figure*}

We collect tactile and visual data from the tire's contact with 12 different terrains to validate the effectiveness of bimodal tires and multimodal sensing networks. We capture 150 images for each terrain, as shown in Fig.~\ref{fig:Bimodalexp}. These terrains contain rubber tracks, painted roads, brick roads, lawns, and gravel roads made of different colored and sized stones. To better demonstrate the effectiveness of the system, we design different comparison experiments.

First, to test the effectiveness of the smart tires, we compare the classification accuracy in different modalities. We process and divide the collected dataset into the following sets: 1) Tactile data only (TO): raw data is segmented to focus solely on the tactile region; 2) Visual data only (VO): Raw data is segmented to focus solely on the transparent and visible region; 3) Raw VisuoTactile data (RVT): Raw data encompass both the tactile region and visible region; 4) Segmented VisuoTactile data (SVT): Raw data is segmented into tactile region and visible region. For all cases, we split the dataset into 70\% for training and 30\% for evaluation. To simulate noise caused by mud on the transparent region, we add salt-and-pepper noise to the visual modality. We train our network on an Intel(R) Xeon(R) Gold 5218 with a single GeForce RTX A6000 for 80 epochs. The learning rate is 2e-5, and we repeated the experiment on 3 different seeds. The training results and related indicators are shown in the last row of Table ~\ref{t1} and Fig.~\ref{fig:Bimodalexp}(f). It can be seen from the results that the classification method with bimodal fusion has higher accuracy, and SVT also achieves a better classification performance than RVT.


\begin{table}[]

\caption{Test results under different modal and network conditions }
\begin{center}
\begin{tabular}{m{1.2cm}<{\centering} p{1.4cm}<{\centering} m{1.4cm}<{\centering} m{1.4cm}<{\centering} m{1.4cm}<{\centering}}

\hline
       & \textbf{TO}    & \textbf{VO}              & \textbf{RVT}              & \textbf{SVT} \\ \hline
ResNet~\cite{He2015DeepRL}    & \textbf{\textbf{92.7\%}}/93.4\%    & 59.0\%/60.4\%       & 82.3\%/83.6\%      & 92.4\%/94.0\%     \\
LSTM~\cite{Shi2015ConvolutionalLN}   & 86.3\%/86.7\%     & 74.4\%/77.2\%      & 84.8\%/87.5\%    & 82.4\%/85.7\%      \\
ViT~\cite{dosovitskiy2020image} & 64.2\%/64.9\% & 71.6\%/72.2\% & 95.0\%/95.6\% &  86.4\%/86.9\%  \\
ExViT~\cite{yao2023extended} & 77.6\%/79.0\% & 66.0\%/71.2\% & 90.2\%/91.5\% & 88.5\%/90.1\% \\
{\bf MMVTT} & 92.6\%/\textbf{\textbf{\textbf{\textbf{93.8\%}}}} & \textbf{\textbf{\textbf{\textbf{77.9\%}}}}/\textbf{\textbf{\textbf{\textbf{77.9\%}}}} & \textbf{\textbf{\textbf{\textbf{96.7\%}}}}/\textbf{\textbf{\textbf{\textbf{97.5\%}}}} & \textbf{\textbf{\textbf{\textbf{98.1\%}}}}/\textbf{\textbf{\textbf{\textbf{98.7\%}}}}   \\ \hline
\end{tabular}
\end{center}
\label{t1}
\end{table}

Second, to further validate the effectiveness of our proposed network, we compare it with current classical classification algorithms. Specifically, we benchmark our method against two baselines: 1) ResNet: We utilize a pre-trained ResNet50 to extract global features for each modality. These global features are then concatenated and passed through a classification head; 2) LSTM~\cite{Shi2015ConvolutionalLN}: We employ a ResNet18 to extract patched features, which are subsequently processed by an LSTM to achieve a fused feature. The fused feature is passed through a classification head to produce the final output; 3) ViT~\cite{dosovitskiy2020image}: We use a Vision Transformer as another baseline for encoder comparison, leveraging the same transformer backbone as our method; 4) ExViT~\cite{yao2023extended}: A multimodal classification method that replaces our ResNet encoder with a CNN. Our method, MultiModal VisoTactile Transformer (MMVTT), is conceptually similar to the LSTM approach but replaces the recurrent structure with an attention block. To ensure a fair comparison, we design the classification heads of these networks to be as similar as possible. The parameters of MMVTT, ExVit, ViT, LSTM, and ResNet are approximately 14M, 11M, 14M, 13M, and 18M, respectively. All three networks are trained on the multimodal dataset with a learning rate of 2e-5 for 80 epochs. Additionally, we conduct the experiments using three different random seeds to minimize the likelihood of incidental results.

The training results and related indicators are shown in Table~\ref{t1} and Fig.~\ref{fig:Bimodalexp}. From the results, we can see that our proposed network outperforms all baselines (Resnet, LSTM, ViT and ExViT) in both last-10-epoch-average and maximum classification accuracy in most cases, especially in bimodal classification, proving our proposed network's effectiveness in dealing with multimodal data.


\subsection{Multimodal Terrain Classification (VTire Bimodal Data + External Visual Data)}

\begin{figure*}
	\centering
  \includegraphics[width=0.95\textwidth]{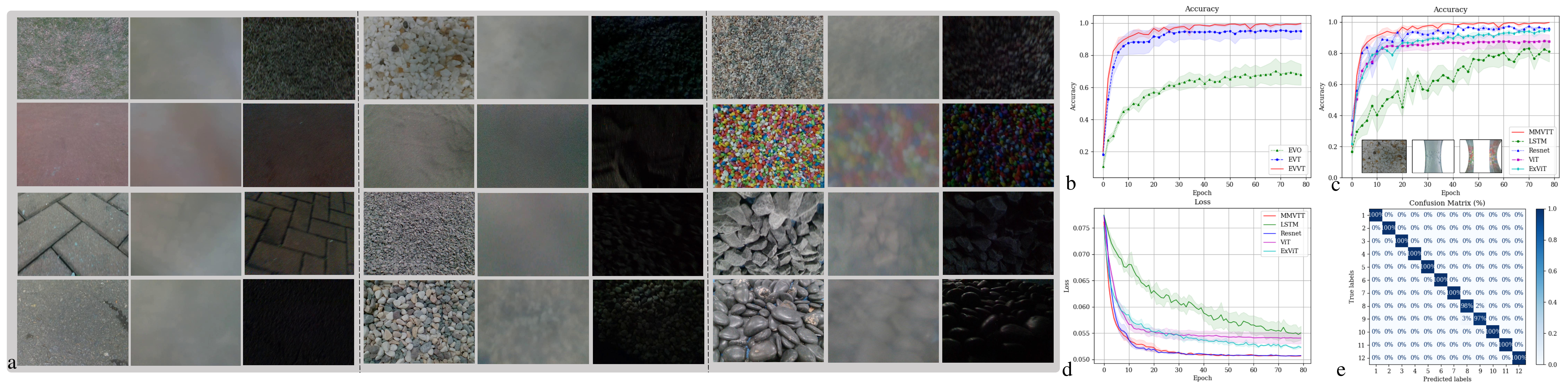}
	\caption{ Multimodal terrain classification. (a) Visual images detected by an external camera under sunny, smoky, and dark conditions; (b) The classification result of our proposed network in different modalities (EVO: external visual only, EVT: external visual data + tactile data,	EVVT: external visual data + segmented visual data + tactile data.); (c) The classification result of different networks in segmented visuotactile data with EVVT; (d) The loss of different networks in segmented visuotactile data with EVVT; (e) The confusion matrix of our proposed network for EVVT.}  \label{fig:multimodalexp}
\end{figure*}

The most common method for terrain recognition is visual processing because the visual information has a greater detection distance and range. Still, for some smoke, darkness, and other scenes, the visual detection effect will receive a great impact, but the tactile information has better stability. Therefore, we think that the visual-tactile fusion approach can solve the problem of terrain perception in complex scenes.
To prove this, we collect visual information from 12 different scenes of normal, smoke, and darkness and compare it with the effect of terrain classification of smart tires. Among them, the data for the smoke scene was collected using a lens wrapped in a semi-transparent film (0.5mm thickness Expanded polyethylene (EPE) material). We capture 150 images for each terrain, as shown in Fig.~\ref{fig:multimodalexp}(a).

In this experiment, we consider three modalities of inputs: 1) External Visual Only (EVO): Only   external vision is used for classification (Although the smart tire itself has visual perception capabilities, its perception is often fuzzy with a limited viewing angle. Therefore, we consider adding external vision to enhance detection accuracy further.);  2) External Visual data + Tactile data (EVT): Both external vision and segmented tactile region data are employed; 3) External Visual data + segmented VisuoTactile data (EVVT): All available modalities are utilized. The multimodal data was randomly split into training and validation datasets with a 7:3 ratio. To assess the capability of each modality, we employed MMVTT across the three input modes mentioned, as MMVTT has demonstrated superior performance, thereby mitigating potential assessment bias that could arise from the use of less effective models. Each configuration was run with 3 random initializations for 80 epochs, using a learning rate of 2e-5. As illustrated in the last row of Table~\ref{t2} and Fig.~\ref{fig:multimodalexp}(b),  the results demonstrate that the EVVT configuration achieves the highest last-10-epoch-average and maximum accuracy with efficient modality fusion.

Furthermore, we compare our method with the previously mentioned baselines: ResNet, LSTM, ViT and ExViT. We utilized all modalities as input, corresponding to the EVVT input mode, and split the data into a training ratio of 0.7. The learning rate is set to 2e-5, and the training process is repeated three times. Table~\ref{t2} and Fig.~\ref{fig:multimodalexp} present the training and evaluation results. Notably, our network outperforms the other baselines and achieves an accuracy exceeding 99\%.  To test the classification of tires at different weights, we test five different loads (0 kg,10 kg,20 kg,30 kg,35 kg) on three different terrains.  We tested the data using a classification network, and the classification effect is the same compared to the previous results.

\begin{table}[]
\caption{Test results under different modal and network conditions }
\begin{center}
\begin{tabular}{m{1.4cm}<{\centering} p{1.8cm}<{\centering} m{1.8cm}<{\centering} m{1.8cm}<{\centering}}
\hline
    & \textbf{EVO}   & \textbf{EVT}   & \textbf{EVVT} \\ \hline
 ResNet   & -     & -     & 96.1\%/97.5\%   \\
  LSTM   & -       & -   & 79.5\%/83.0\%   \\
  ViT    & -   &  -   &  87.4\%/87.7\% \\
 ExViT & - & - & 93.6\%/94.8\% \\
   {\bf MMVTT}  & 68.5\%/70.2\%              & 95.1\%/95.6\% & \textbf{\textbf{\textbf{\textbf{99.2\%}}}}/\textbf{\textbf{\textbf{\textbf{99.7}}}}\% \\ \hline
\end{tabular}
\end{center}
\label{t2}
\end{table}

\subsection{Object Search Experiment on the Ground}

{
Finding objects dropping on the ground is a big pain point for humans, especially for transparent objects. Because transparent objects have special optical properties, not only do they lack texture, but their color changes with the background.
Compared with vision, although touch has a smaller detection area, it can get more stable contact information by touching objects (independent of background, and optical properties). However, VTire can solve this problem well with its powerful tactile perception ability. By combining  VTire with sweeping robots and wheeled robots, they can detect the cleanliness of the floor and the presence of foreign objects as they move. }

\begin{figure}
	\centering
  \includegraphics[width=0.45\textwidth]{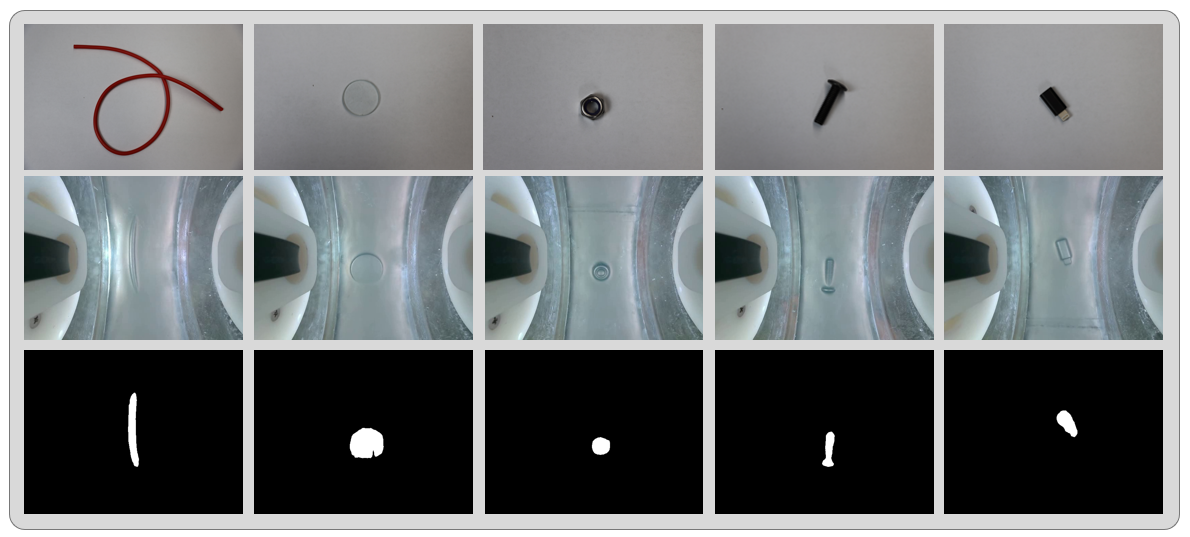}
	\caption{Segmentation results for contacting objects. From left to right: wire, pane, nut, screwdriver, USB converter.  } \label{fig:test2}
\end{figure}

We design an object search experiment to verify the function of VTire on search. First, we collect data from five different objects, namely, rope, lens, nut, screw, and USB converter. A total of 150 images of tires in contact with objects are collected and then manually annotated. After completing the design of the dataset, we train the FCN. After 60 rounds of training, the segmentation accuracy can reach 99\%, and the object search results are shown in Fig.~\ref{fig:test2}.

In addition, we also conduct object search experiments in real scenarios, where objects are randomly thrown on the ground and searched using VTires. After 50 experimental tests, the success rate of finding the object successfully is 98\%, indicating the potential of VTires in the field of searching.

\subsection{Cracks Detection Experiment}

\begin{figure}
	\centering
  \includegraphics[width=0.45\textwidth]{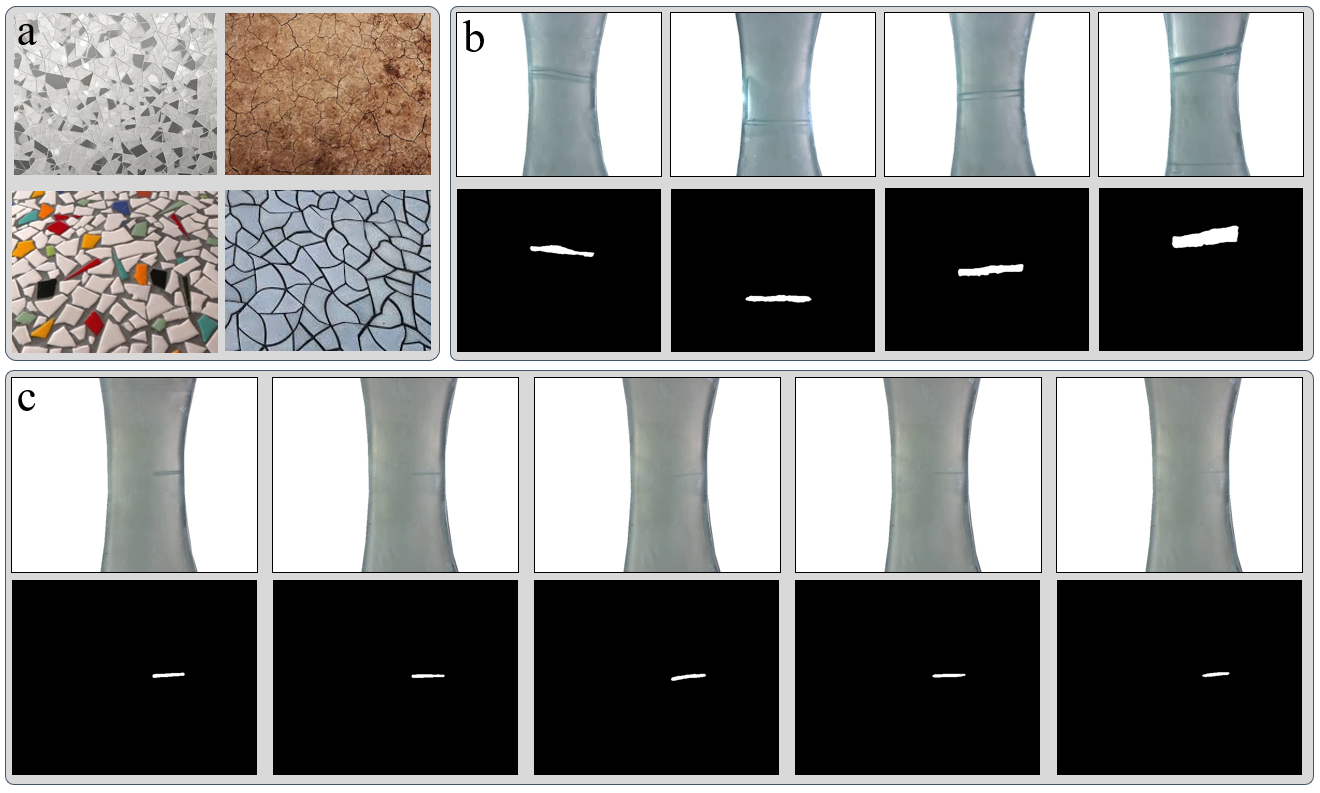}
	\caption{ Ground cracked detection results. (a) Floor tiles with a cracked decorative pattern; (b) Segmentation results in broken ground. (c) Perception effect of needles of different thicknesses. From left to right: 0.5 mm; 0.4 mm; 0.3 mm; 0.25 mm; 0.2 mm.  } \label{fig:cracked}
\end{figure}

 Besides object searching, crack searching can be a very rewarding endeavor. Floor tiles with a variety of textures are often integrated with cracks, which can cause significant interference with visual detection\cite{hu20243d,tang2024obstacle}, as shown in Fig.~\ref{fig:cracked}(a), while tactile perception can avoid the impact of the pattern on detection. Combining home robots with VTire can be used to inspect flooring equipment in real-time, in the renovation industry to check the quality of floor coverings, and in transportation to assess road strength, integrity, and other indicators. 

To test the effectiveness of VTire on crack detection, we design a crack detection experiment. First, we collect fragments of different sizes of cracks. Then, we capture 120 images of tires in contact with cracks. After that, we annotate the image at the pixel level, and unlike object searching, we mask the visuotactile part of the tires to prevent other regions from influencing the detection of the tactile region, as crack detection is much more difficult. Finally, we train the FCN using the dataset, and the results obtained are shown in Fig.~\ref{fig:cracked}; after 60 rounds of training, the training accuracy can reach 98\%, and the cracks search results are shown in Fig.~\ref{fig:cracked}(b). In addition, we use needles of different thicknesses to quantify the tactile resolution of the tires, which is tested to 0.2 mm, as shown in Fig.~\ref{fig:cracked}(c).

\subsection{Tire Damage  Detection Experiment}

Tires are susceptible to damage due to contact with sharp objects or long-distance driving while the vehicle is driving. Compared with traditional smart tires, VTire can detect damage in real-time, such as crack, abrasion, and nail penetration, thanks to high-resolution tactile data.

\begin{figure}
	\centering
  \includegraphics[width=0.45\textwidth]{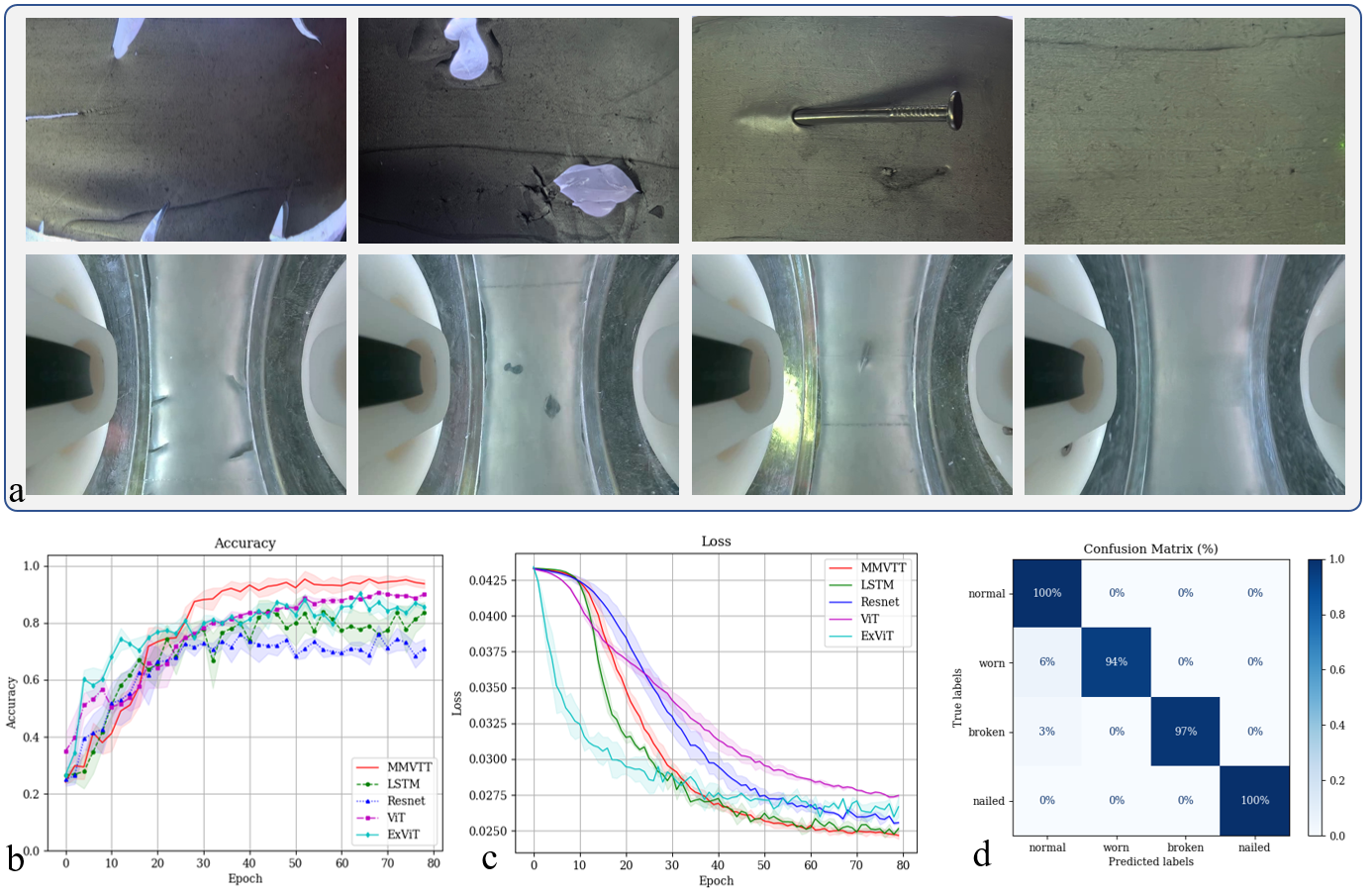}
	\caption{Damage detection. (a) Common tire damage: cracks, irregular wear, punctures, and normal tires (from left to right); (b) Classification accuracy of different networks; (c) Loss curve of different networks; (d) Confusion matrix for the classification result of our proposed network. } \label{fig:breakage}
\end{figure}

We experiment with different kinds of discrimination to evaluate the capability of detecting damage. We consider 3 types of damage: 1) cracks, 2) irregular wear, 3) punctures, and a normal state. We collect 120 images for each state of tile and split 70\% for training. We also add pepper and salt noise to mimic the possible effects of the real environment. MMVTT, ExViT, ViT, LSTM, and ResNet are employed to learn from the training data. We choose a learning rate of 2e-5 and run the experiment on 3 random seeds for 80 epochs. Fig.~\ref{fig:breakage} illustrates the training results and related metrics. We find that MMVTT can detect the damage accurately and correctly classify the type of damage in more than 97\% of cases.

\subsection{Load Weight Perception Experiment}

\begin{figure}
	\centering
  \includegraphics[width=0.45\textwidth]{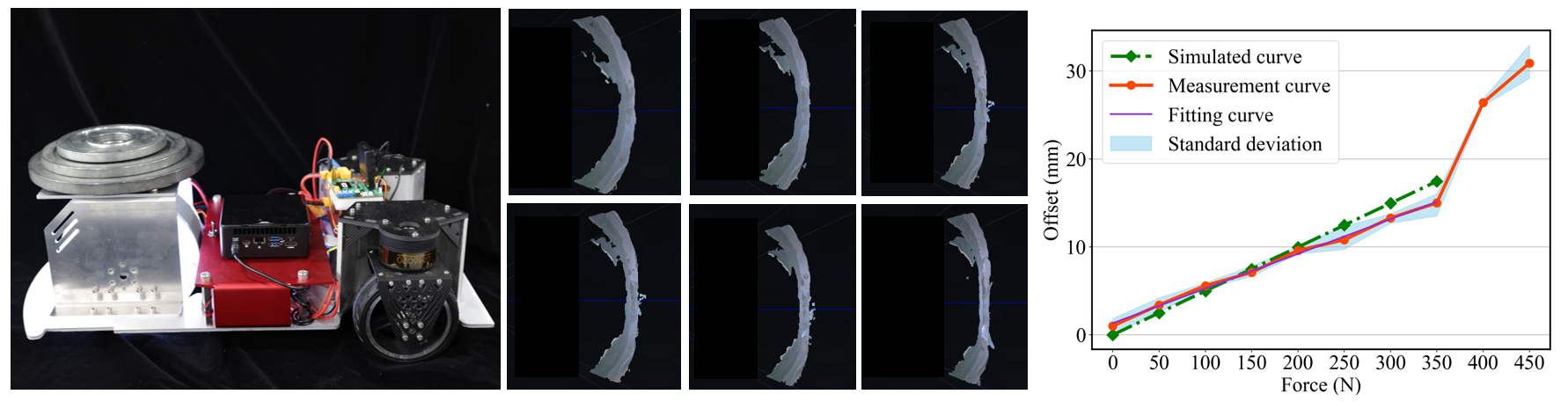}
	\caption{ Load weight perception experiment. (Left) Weight test scenario; (Middle) Depth information of tires under different loads; (Right) Corresponding curve between offset and load.
} \label{fig:weighttest}
\end{figure}

To evaluate the load weight perception performance of the tires, we build a test rig, as shown in Fig.~\ref{fig:weighttest}(left), where we use dumbbell pieces as loads to test the tires' load capacity. 
After obtaining the load limits of the tires, we conduct load capacity perception experiments on the tires. 
Based on the force sensing algorithm proposed above, we fit the curve between force and offset using the equation. From Fig.~\ref{fig:weighttest}(right), we can see that the load is linearly related to the offset when the tire is 0 $\sim$ 35 kg, and the results are very close to the results of  FEA. We test 10 weights, each with 5 measurements, with a weight perception accuracy of 0.75 kg. After exceeding 35 kg, the tire will undergo a sudden change due to breaking the load limit of the tire.

\begin{figure}
	\centering
  \includegraphics[width=0.46\textwidth]{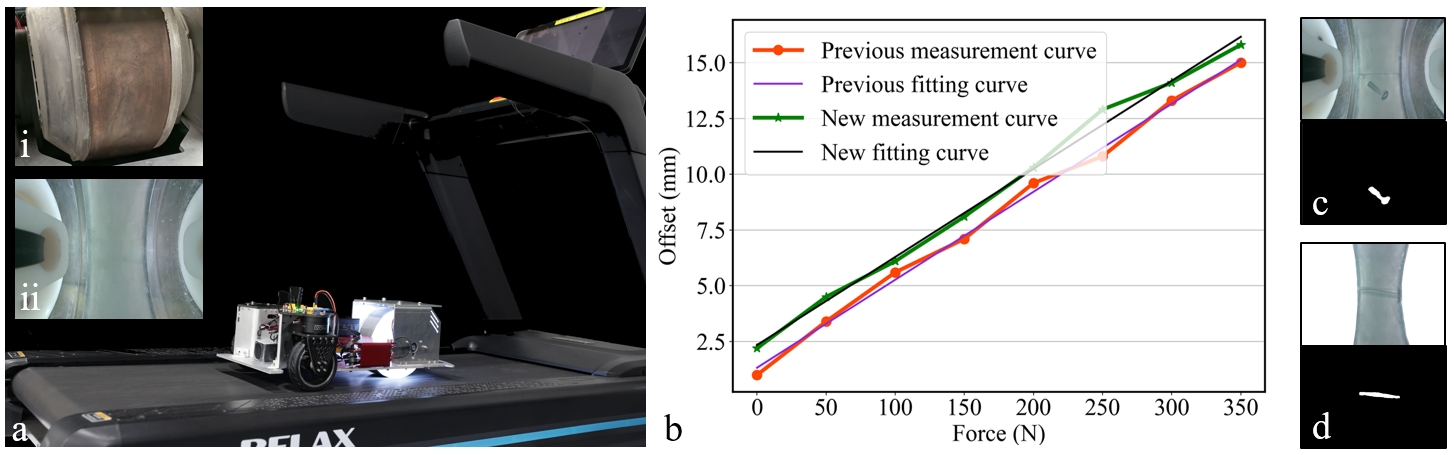}
	\caption{  Performance test experiment. (a) Durability test platform. (i) Tire appearance after testing; (ii) Internal image after testing. (b) Load test data after durability test.  (c) Object search data after durability test; (d) Crack detection data after durability test. } \label{fig:Durability}
\end{figure}

\begin{figure}[ht]
	\centering
  \includegraphics[width=0.45\textwidth]{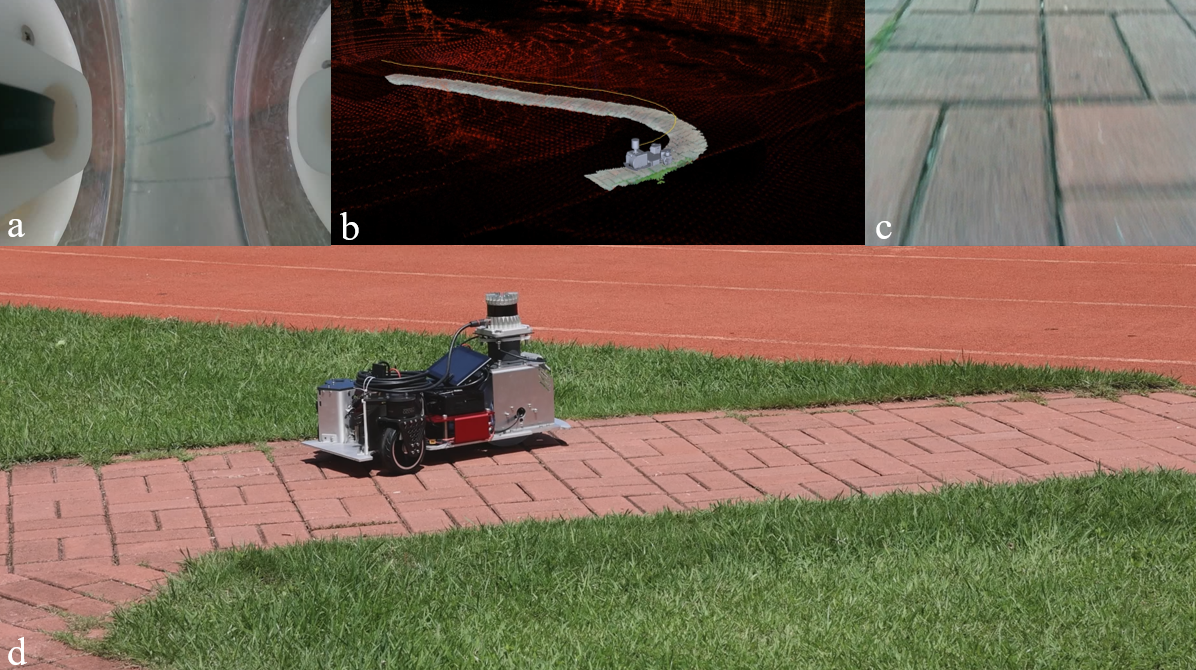}
	\caption{Outdoor scene test experiment. (a) Visuotactile image; (b) Camera-LiDAR mapping result; (c) External visual image; (d) Test scene. } \label{fig:Outdoor}
\end{figure}

\subsection{Durability Test}

Durability is an important indicator of a tire. However, the special properties of a specialty tire often limit its durability; for example, street car tires can last about 15,000 km, but the life of an F1 tire is between 60 and 120 km\cite{Durability,weissman2003extending}. 
To test the VTire's durability, we set up a simple test system on a treadmill. The system is driven continuously for 200 km at a speed of 10 km/h, as shown in Fig.~\ref{fig:Durability}. After the durability test, we re-evaluated the tire's performance in terrain classification, object search, crack detection, and load sensing experiments to verify its stability and durability.

Specifically, after the 200 km durability test, the VTire maintained consistent performance in all key metrics. The terrain classification accuracy can up to 98\%, the object search experiment still provided clear contour information, and the crack detection resolution was stable at 0.2 mm. For load sensing, although prolonged use caused a slight shift in the fitting curve, recalibration restored the detection accuracy to its original level. These results confirm that the VTire retains high performance even after extended use.  In addition, the tire adopts a modular design so that when the sensing skin is damaged, only the outermost sensing layer needs to be replaced, not the entire tire, which reduces the cost of tire maintenance.

\subsection{Outdoor Scene Test}

Besides the indoor scene, we test it in an outdoor scene, as shown in Fig.~\ref{fig:Outdoor}. 
We implement mapping using the RealSense D435i fused with LiDAR and employ the multimodal classification algorithm to detect the terrains. We test terrains such as grass, road, tartan track, and brick, achieving good detection and mapping results. (Details can be referred to the video file.)

\section{Conclusion}
This paper proposes a bimodal smart visuotactile trie named VTire, which can obtain large-area and high-resolution tactile \& visual data. We optimize the material and structure to obtain highly elastic, durable, and transparent tires and propose a complete set of preparation processes. The tires can withstand loads exceeding 35 kg and have a durability of over 200 km. Algorithmically, we propose a transformer-based multimodal classification algorithm, a load detection based on finite element analysis, and a contact segmentation algorithm. In addition, to validate the algorithm's performance, we build a mobile platform and propose a set of visual and tactile datasets in different terrain and visibility situations. After experimental validation, the accuracy of our proposed multimodal classification method can reach 99.2\%  in terrain classification, 98\% success rate in object search, and 0.75 kg accuracy in weight sensing, which proves that VTire has a high application value in terrain sensing and object searching scenarios.

The tire also has some limitations. Due to our single-camera solution, the tire is unsuitable for high-speed scenes.   Future work will focus on integrating advanced vision systems, applying the tire to wheeled robots, and exploring new materials, real-time algorithms, and autonomous robot integration for complex terrains.

\bibliographystyle{ieeetr}
\bibliography{ref}

\begin{thebibliography}{10}

\bibitem{10646374}
B.~Yang, Q.~Sun, R.~Fu, C.~Wang, Y.~Guo, and L.~Zhou, ``A model predictive control-based electronic differential control strategy for distributed-drive buses considering the reduction of tire wear,'' {\em IEEE/ASME Transactions on Mechatronics}, pp.~1--12, 2024.

\bibitem{10597662}
Y.~Wang, T.~Chen, X.~Rong, G.~Zhang, Y.~Li, and Y.~Xin, ``Design and control of skater: A wheeled-bipedal robot with high-speed turning robustness and terrain adaptability,'' {\em IEEE/ASME Transactions on Mechatronics}, pp.~1--12, 2024.

\bibitem{lee2017intelligent}
H.~Lee and S.~Taheri, ``Intelligent tires? a review of tire characterization literature,'' {\em IEEE Intelligent Transportation Systems Magazine}, vol.~9, no.~2, pp.~114--135, 2017.

\bibitem{xu2020tire}
N.~Xu, H.~Askari, Y.~Huang, J.~Zhou, and A.~Khajepour, ``Tire force estimation in intelligent tires using machine learning,'' {\em IEEE Transactions on Intelligent Transportation Systems}, vol.~23, no.~4, pp.~3565--3574, 2020.

\bibitem{maurya20203d}
D.~Maurya, S.~Khaleghian, R.~Sriramdas, P.~Kumar, R.~A. Kishore, M.~G. Kang, V.~Kumar, H.-C. Song, S.-Y. Lee, Y.~Yan, {\em et~al.}, ``3d printed graphene-based self-powered strain sensors for smart tires in autonomous vehicles,'' {\em Nature Communications}, vol.~11, no.~1, p.~5392, 2020.

\bibitem{tuononen2011laser}
A.~J. Tuononen, ``Laser triangulation to measure the carcass deflections of a rolling tire,'' {\em Measurement Science and Technology}, vol.~22, no.~12, p.~125304, 2011.

\bibitem{tuononen2009board}
A.~Tuononen, ``On-board estimation of dynamic tyre forces from optically measured tyre carcass deflections,'' {\em International Journal of Heavy Vehicle Systems}, vol.~16, no.~3, pp.~362--378, 2009.

\bibitem{matsuzaki2010optical}
R.~Matsuzaki, N.~Hiraoka, A.~Todoroki, and Y.~Mizutani, ``Optical 3d deformation measurement utilizing non-planar surface for the development of an “intelligent tire”,'' {\em Journal of Solid Mechanics and Materials Engineering}, vol.~4, no.~4, pp.~520--532, 2010.

\bibitem{abad2020visuotactile}
A.~C. Abad and A.~Ranasinghe, ``Visuotactile sensors with emphasis on {Gelsight} sensor: A review,'' {\em IEEE Sensors Journal}, vol.~20, no.~14, pp.~7628--7638, 2020.

\bibitem{lu2024dexitac}
C.~Lu, K.~Tang, M.~Yang, T.~Yue, H.~Li, and N.~F. Lepora, ``Dexitac: Soft dexterous tactile gripping,'' {\em IEEE/ASME Transactions on Mechatronics}, 2024.

\bibitem{10358360}
S.~Cui, S.~Wang, C.~Zhang, R.~Wang, B.~Zhang, S.~Zhang, and Y.~Wang, ``Gelstereo biotip: Self-calibrating bionic fingertip visuotactile sensor for robotic manipulation,'' {\em IEEE/ASME Transactions on Mechatronics}, pp.~1--12, 2023.

\bibitem{singh2015estimation}
K.~B. Singh and S.~Taheri, ``Estimation of tire--road friction coefficient and its application in chassis control systems,'' {\em Systems Science \& Control Engineering}, vol.~3, no.~1, pp.~39--61, 2015.

\bibitem{barbosa2021lateral}
B.~H.~G. Barbosa, N.~Xu, H.~Askari, and A.~Khajepour, ``Lateral force prediction using gaussian process regression for intelligent tire systems,'' {\em IEEE Transactions on Systems, Man, and Cybernetics: Systems}, vol.~52, no.~8, pp.~5332--5343, 2021.

\bibitem{oh2012development}
H.~Oh, K.~Lee, K.~Eun, S.-H. Choa, and S.~S. Yang, ``Development of a high-sensitivity strain measurement system based on a sh saw sensor,'' {\em Journal of Micromechanics and Microengineering}, vol.~22, no.~2, p.~025002, 2012.

\bibitem{zhang2004design}
X.~Zhang, Z.~Wang, L.~Gai, Y.~Ai, and F.~Wang, ``Design considerations on intelligent tires utilizing wireless passive surface acoustic wave sensors,'' in {\em Fifth World Congress on Intelligent Control and Automation (IEEE Cat. No. 04EX788)}, vol.~4, pp.~3696--3700, 2004.

\bibitem{4453936}
J.~Yi, ``A piezo-sensor-based “smart tire” system for mobile robots and vehicles,'' {\em IEEE/ASME Transactions on Mechatronics}, vol.~13, no.~1, pp.~95--103, 2008.

\bibitem{10552090}
X.~Sun, Z.~Quan, Y.~Cai, L.~Chen, and B.~Li, ``Direct tire slip angle estimation using intelligent tire equipped with pvdf sensors,'' {\em IEEE/ASME Transactions on Mechatronics}, pp.~1--11, 2024.

\bibitem{mendoza2020strain}
M.~F. Mendoza-Petit, D.~Garc{\'\i}a-Pozuelo, V.~D{\'\i}az, and O.~Olatunbosun, ``A strain-based intelligent tire to detect contact patch features for complex maneuvers,'' {\em Sensors}, vol.~20, no.~6, p.~1750, 2020.

\bibitem{yunta2019influence}
J.~Yunta, D.~Garcia-Pozuelo, V.~Diaz, and O.~Olatunbosun, ``Influence of camber angle on tire tread behavior by an on-board strain-based system for intelligent tires,'' {\em Measurement}, vol.~145, pp.~631--639, 2019.

\bibitem{gubaidullin2019microwave}
R.~Gubaidullin, T.~Agliullin, O.~Morozov, A.~Z. Sahabutdinov, and V.~Ivanov, ``Microwave-photonic sensory tire control system based on fbg,'' in {\em 2019 Systems of Signals Generating and Processing in the Field of on Board Communications}, pp.~1--6, 2019.

\bibitem{longoria2019wheel}
R.~G. Longoria, R.~Brushaber, and A.~Simms, ``An in-wheel sensor for monitoring tire-terrain interaction: Development and laboratory testing,'' {\em Journal of Terramechanics}, vol.~82, pp.~43--52, 2019.

\bibitem{10354894}
L.~Hu, F.~Xue, C.~Yao, Y.~Li, J.~Wei, P.~Wang, Z.~Zhu, and Z.~Jia, ``Terrain classification using inside-wheel cameras based on wheel-terrain interaction characteristics,'' in {\em 2023 IEEE International Conference on Robotics and Biomimetics (ROBIO)}, pp.~1--6, 2023.

\bibitem{kim2020road}
H.-J. Kim, J.-Y. Han, S.~Lee, J.-R. Kwag, M.-G. Kuk, I.-H. Han, and M.-H. Kim, ``A road condition classification algorithm for a tire acceleration sensor using an artificial neural network,'' {\em Electronics}, vol.~9, no.~3, p.~404, 2020.

\bibitem{khaleghian2017terrain}
S.~Khaleghian and S.~Taheri, ``Terrain classification using intelligent tire,'' {\em Journal of Terramechanics}, vol.~71, pp.~15--24, 2017.

\bibitem{eun2016highly}
K.~Eun, K.~J. Lee, K.~K. Lee, S.~S. Yang, and S.-H. Choa, ``Highly sensitive surface acoustic wave strain sensor for the measurement of tire deformation,'' {\em International Journal of Precision Engineering and Manufacturing}, vol.~17, pp.~699--707, 2016.

\bibitem{9882387}
J.~Jiang, G.~Cao, A.~Butterworth, T.-T. Do, and S.~Luo, ``Where shall i touch? vision-guided tactile poking for transparent object grasping,'' {\em IEEE/ASME Transactions on Mechatronics}, vol.~28, no.~1, pp.~233--244, 2023.

\bibitem{vaswani2017attention}
A.~Vaswani, N.~Shazeer, N.~Parmar, J.~Uszkoreit, L.~Jones, A.~N. Gomez, {\L}.~Kaiser, and I.~Polosukhin, ``Attention is all you need,'' {\em Advances in Neural Information Processing Systems}, vol.~30, 2017.

\bibitem{Ba2016LayerN}
J.~Ba, J.~R. Kiros, and G.~E. Hinton, ``Layer normalization,'' {\em ArXiv}, vol.~abs/1607.06450, 2016.

\bibitem{Chen2022VisuoTactileTF}
Y.~Chen, A.~Sipos, M.~V. der Merwe, and N.~Fazeli, ``Visuo-tactile transformers for manipulation,'' in {\em Conference on Robot Learning}, 2022.

\bibitem{He2015DeepRL}
K.~He, X.~Zhang, S.~Ren, and J.~Sun, ``Deep residual learning for image recognition,'' {\em 2016 IEEE Conference on Computer Vision and Pattern Recognition (CVPR)}, pp.~770--778, 2015.

\bibitem{long2015fully}
J.~Long, E.~Shelhamer, and T.~Darrell, ``Fully convolutional networks for semantic segmentation,'' in {\em IEEE Conference on Computer Vision and Pattern Recognition (CVPR)}, pp.~3431--3440, 2015.

\bibitem{Shi2015ConvolutionalLN}
X.~Shi, Z.~Chen, H.~Wang, D.~Y. Yeung, W.-K. Wong, and W.~chun Woo, ``Convolutional {LSTM} network: A machine learning approach for precipitation nowcasting,'' in {\em Neural Information Processing Systems}, 2015.

\bibitem{dosovitskiy2020image}
A.~Dosovitskiy, L.~Beyer, A.~Kolesnikov, D.~Weissenborn, X.~Zhai, T.~Unterthiner, M.~Dehghani, M.~Minderer, G.~Heigold, S.~Gelly, {\em et~al.}, ``An image is worth 16x16 words: Transformers for image recognition at scale,'' {\em arXiv preprint arXiv:2010.11929}, 2020.

\bibitem{yao2023extended}
J.~Yao, B.~Zhang, C.~Li, D.~Hong, and J.~Chanussot, ``Extended vision transformer (exvit) for land use and land cover classification: A multimodal deep learning framework,'' {\em IEEE Transactions on Geoscience and Remote Sensing}, vol.~61, pp.~1--15, 2023.

\bibitem{hu20243d}
K.~Hu, Z.~Chen, H.~Kang, and Y.~Tang, ``3d vision technologies for a self-developed structural external crack damage recognition robot,'' {\em Automation in Construction}, vol.~159, p.~105262, 2024.

\bibitem{tang2024obstacle}
Y.~Tang, S.~Qi, L.~Zhu, X.~Zhuo, Y.~Zhang, and F.~Meng, ``Obstacle avoidance motion in mobile robotics,'' {\em Journal of System Simulation}, vol.~36, no.~1, pp.~1--26, 2024.

\bibitem{Durability}
``Durability of tires.'' https://www.deccanherald.com/sports/f1-racing/lasting-little-60km-tyres-are-716735.html.

\bibitem{weissman2003extending}
S.~L. Weissman, J.~L. Sackman, D.~Gillen, and C.~Monismith, ``Extending the lifespan of tires,'' {\em Sympletic Engineering Corporation. Institute for Transportation Studies. University of California at Berkeley}, 2003.

\end{thebibliography}

\begin{IEEEbiography}[{\includegraphics[width=1in,height=1.25in,clip,keepaspectratio]{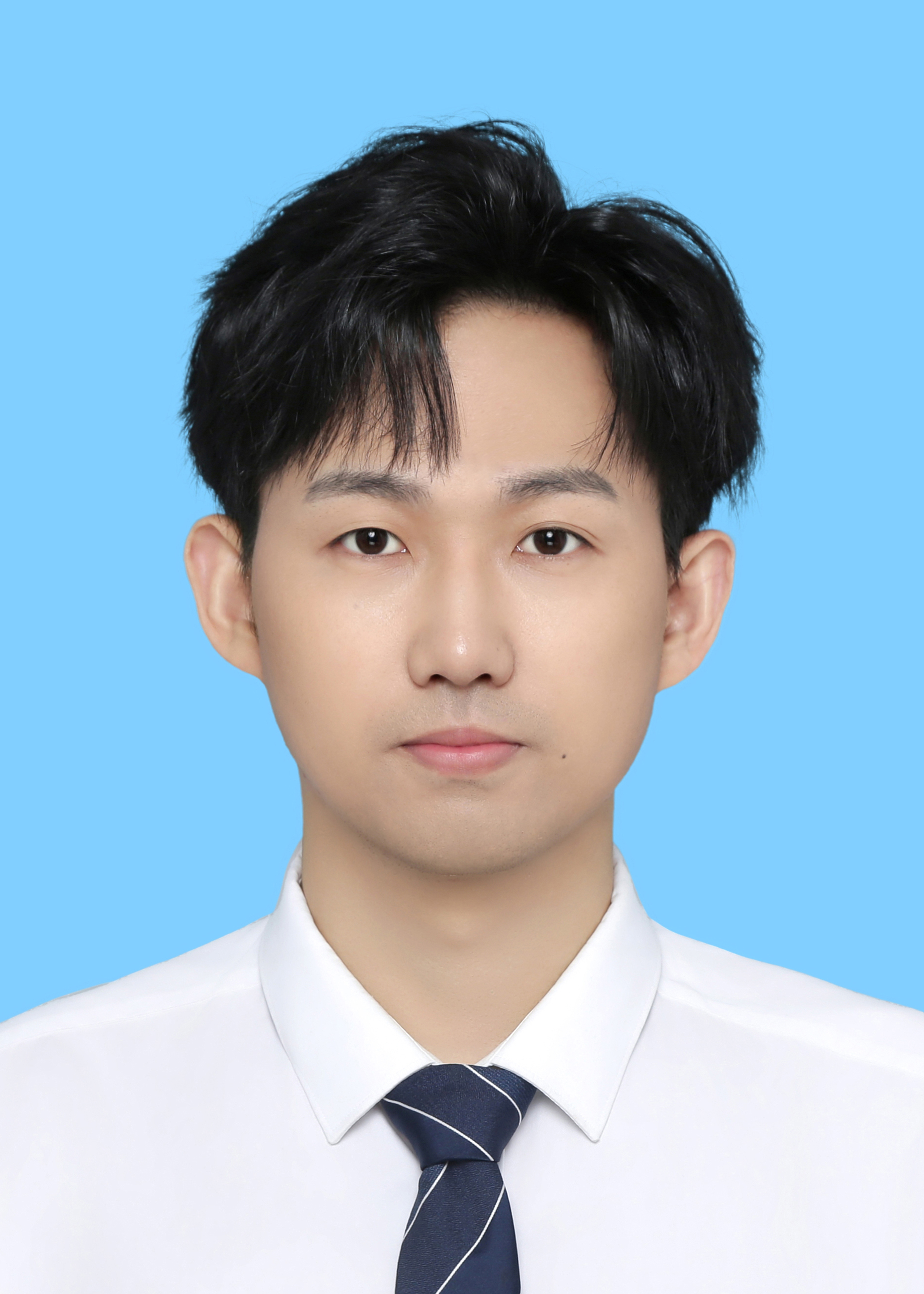}}]{Shoujie Li}
received the B.Eng. degree in electronic information engineering from the College of Oceanography and Space Informatics, China University of Petroleum,
Tsingtao, China, in 2020. He is currently pursuing toward Ph.D. degree in Tsinghua-Berkeley Shenzhen Institute, Shenzhen International Graduate School, Tsinghua University, Shenzhen, China.

His research interests include tactile perception, grasping, and machine learning.
\end{IEEEbiography}

\begin{IEEEbiography}[{\includegraphics[width=1in,height=1.25in,clip,keepaspectratio]{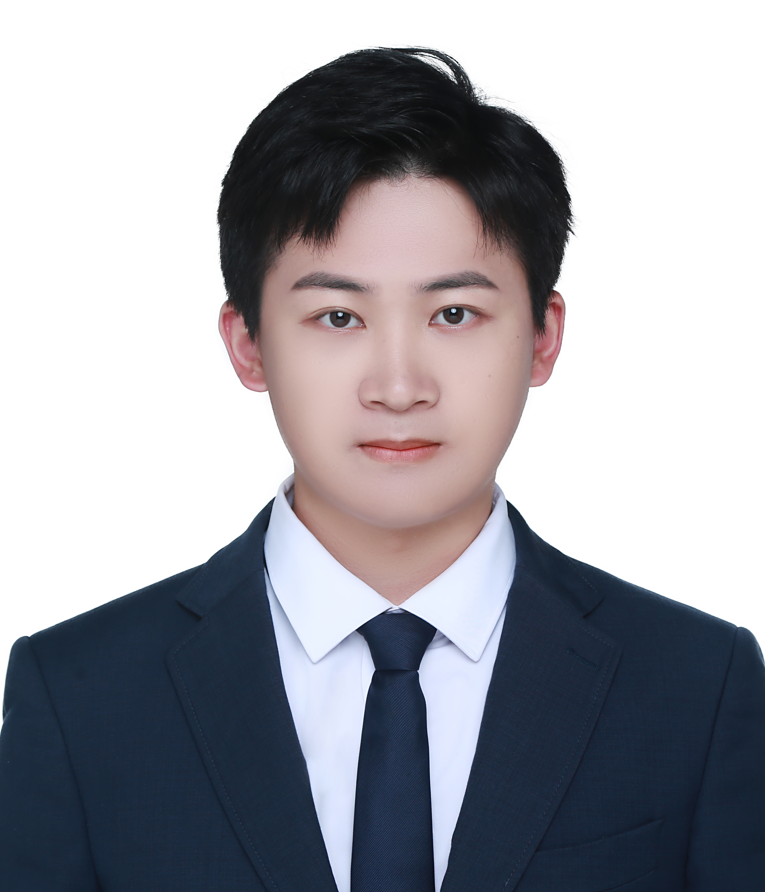}}]{Jianle Xu}
received the B.S. degree in Agricultural Mechanization and Automation from Hainan University, Hainan, China, in 2023. He is currently working toward the M.S.degree in Tsinghua Shenzhen International Graduate School, Tsinghua University, Shenzhen, China.

His research interests include robot dexterous hands and electronic devices.
\end{IEEEbiography}

\begin{IEEEbiography}[{\includegraphics[width=1in,height=1.25in,clip,keepaspectratio]{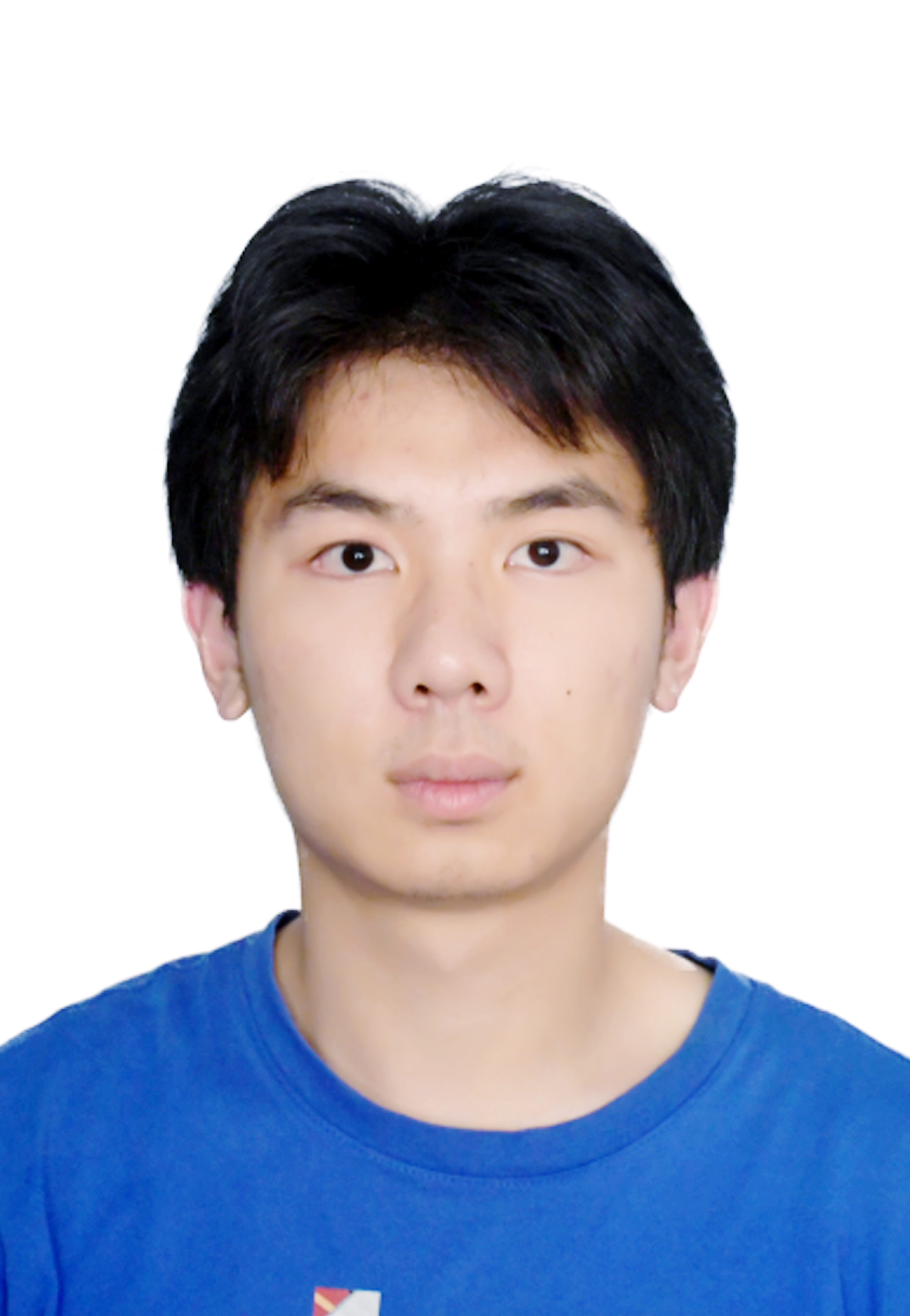}}]{Tong Wu}
received the B.S. degree in electronic engineering from Tsinghua University, Beijing, China, in 2023. He is current working toward the Ph.D. degree in Tsinghua-Berkeley Shenzhen Institute, Tsinghua University, Shenzhen, China.

His research interests include robot manipulation, multimodal sensing, and embodied intelligence.
\end{IEEEbiography}

\begin{IEEEbiography}[{\includegraphics[width=1in,height=1.25in,clip,keepaspectratio]{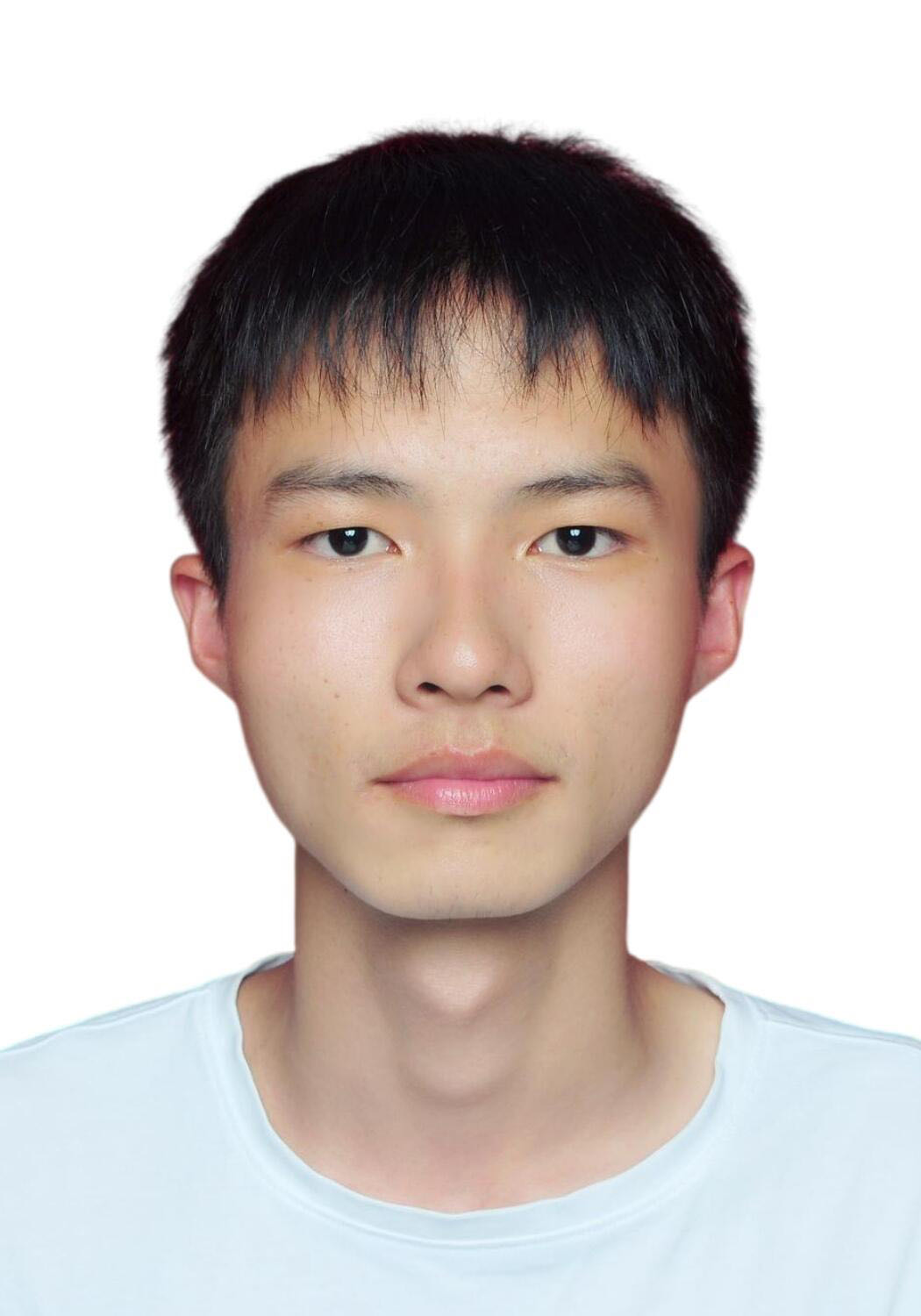}}]{Yang Yang}

is currently an undergraduate student at Sichuan University, Chengdu, China. He will receive his B.S. degree in Engineering Mechanics in 2025. 

His research interests include tactile sensing and robotic dexterous manipulation.
\end{IEEEbiography}

\begin{IEEEbiography}[{\includegraphics[width=1in,height=1.25in,clip,keepaspectratio]{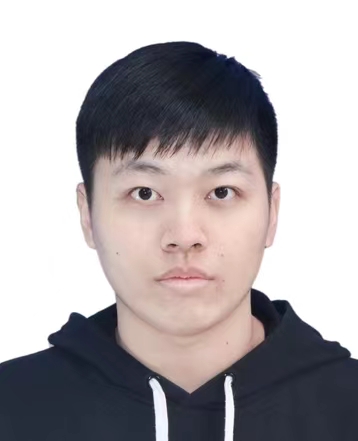}}]{Yanbo Chen}

received the B.S. degree in automation from  Harbin Institute of Technology, Shenzhen, China, in 2023. He is currently working toward the M.S. degree at Tsinghua Shenzhen International Graduate School, Tsinghua University, Shenzhen, China.

His research interests include SLAM and autonomous navigation for robots.
\end{IEEEbiography}

\begin{IEEEbiography}
[{\includegraphics[width=1in,height=1.25in,clip,keepaspectratio]{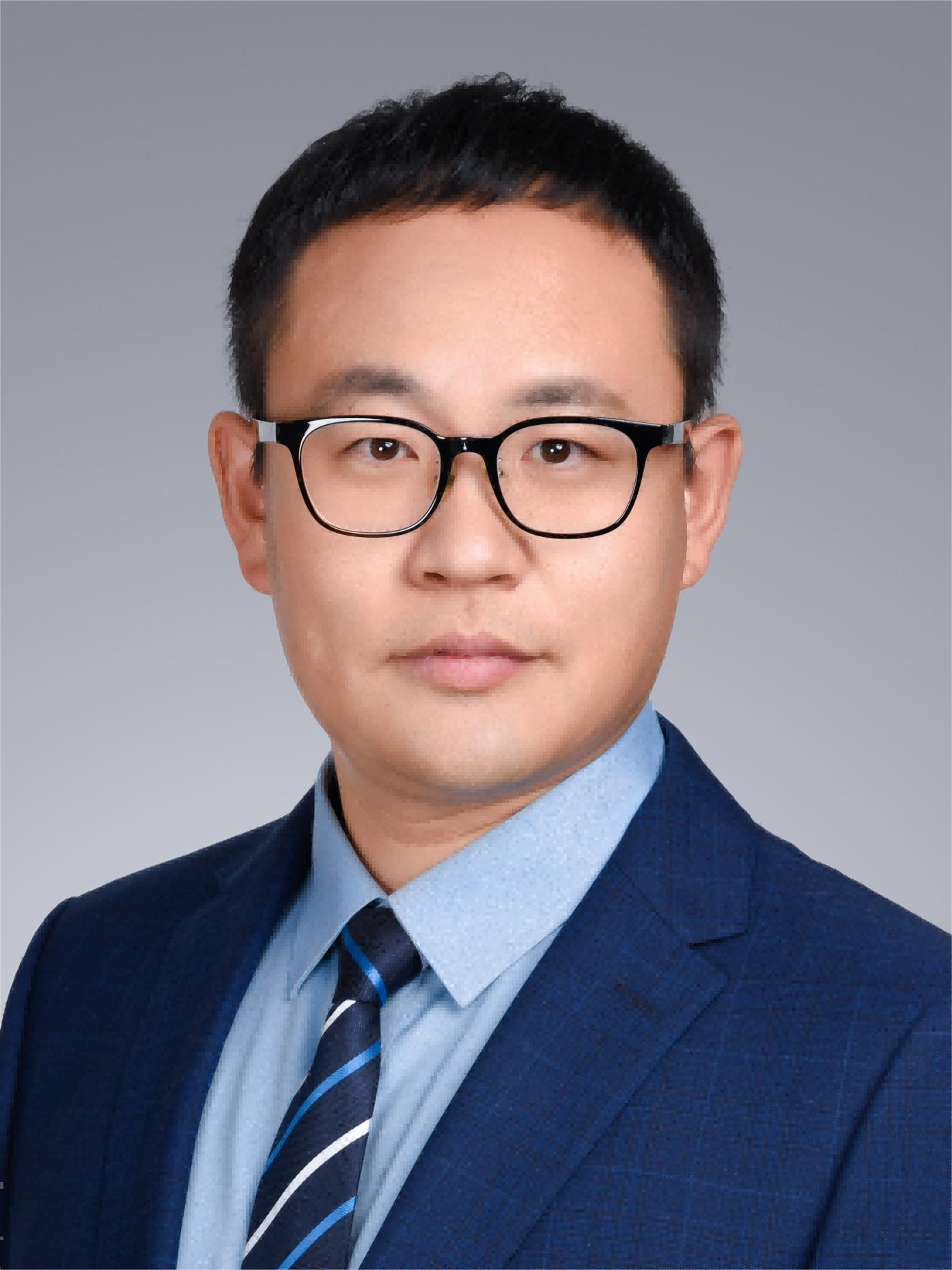}}]{Xueqian Wang}
received the B.E. degree in mechanical design, manufacturing and Automation, Harbin University of Science and Technology, Harbin, China, in 2003, the M.Sc. degree in mechatronic engineering and the  Ph.D. degree in control science and engineering from the Harbin Institute of Technology (HIT), Harbin, China, in 2005 and 2010, respectively. From June 2010 to February 2014, he was the Postdoc Research Fellow with the HIT. He is currently a Professor and the Leader of the Center of Intelligent Control and Telescience, Tsinghua Shenzhen International Graduate School, Tsinghua University, Shenzhen, China. 

His research interests include dynamics modeling, control, and teleoperation of robotic systems.
 
\end{IEEEbiography}
\begin{IEEEbiography}
[{\includegraphics[width=1in,height=1.25in,clip,keepaspectratio]{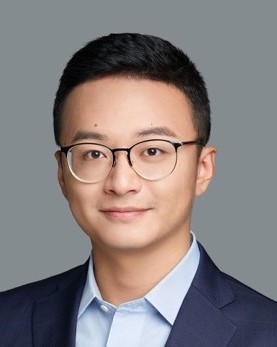}}]{Wenbo Ding}
 received the BS and PhD degrees (Hons.) from Tsinghua University in 2011 and 2016, respectively. He worked as a postdoctoral research fellow at Georgia Tech under the supervision of Professor Z. L. Wang from 2016 to 2019. He is now an associate professor and PhD supervisor at Tsinghua-Berkeley Shenzhen Institute, Institute of Data and Information, Shenzhen International Graduate School, Tsinghua University, where he leads the Smart Sensing and Robotics (SSR) group. He has received many prestigious awards, including the Gold Medal of the 47th International Exhibition of Inventions Geneva and the IEEE Scott Helt Memorial Award.
 
 His research interests are diverse and interdisciplinary, which include self-powered sensors, energy harvesting, and wearable devices for health and robotics with the help of signal processing, machine learning, and mobile computing. 
\end{IEEEbiography}

\begin{IEEEbiography}
[{\includegraphics[width=1in,height=1.25in,clip,keepaspectratio]{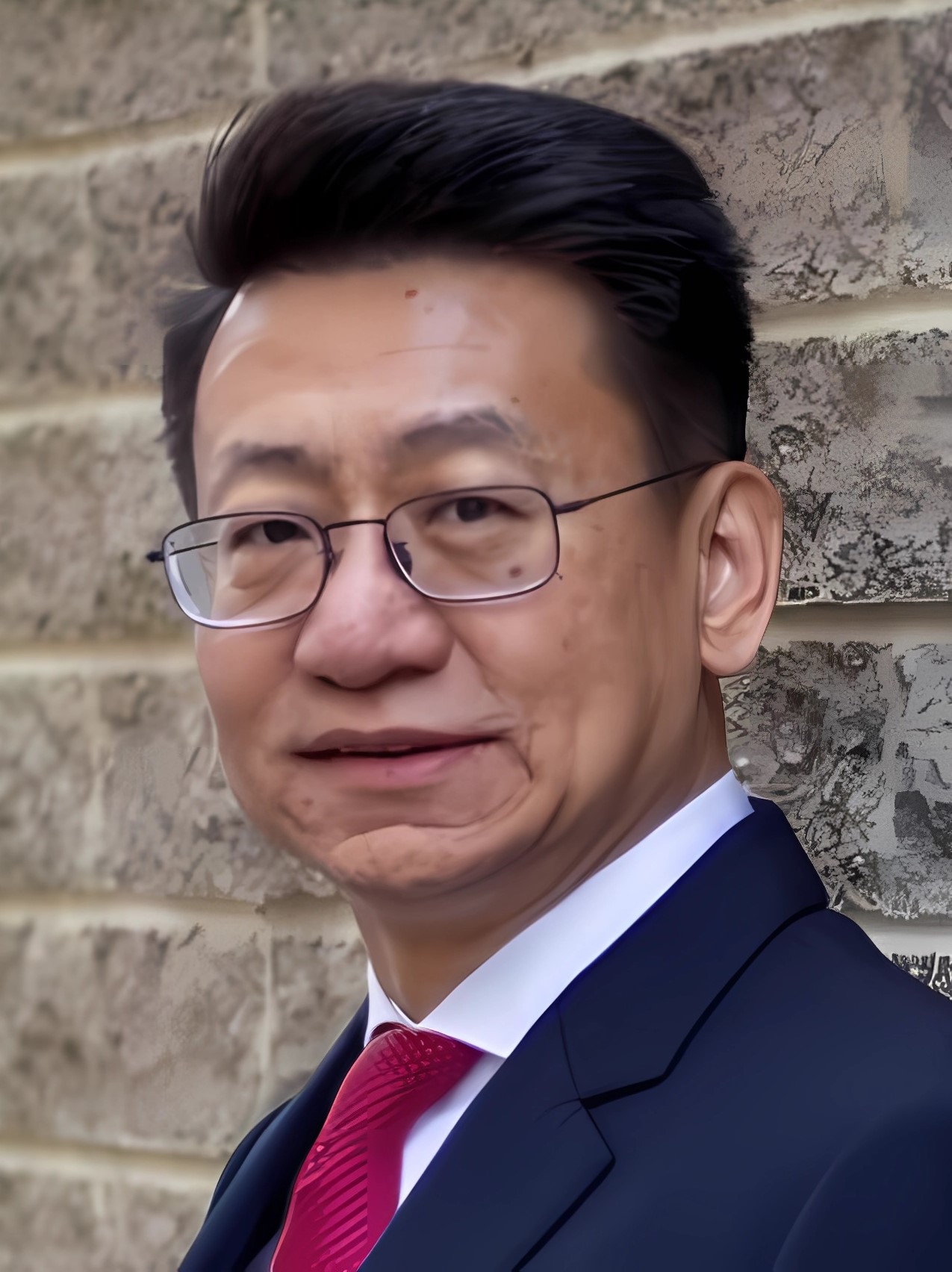}}]{Xiao-Ping Zhang}

 received B.S. and Ph.D. degrees from Tsinghua University, in 1992 and 1996, respectively, both in Electronic Engineering. He holds an MBA in Finance, Economics and Entrepreneurship with Honors from the University of Chicago Booth School of Business, Chicago, IL.
He is Chair Professor at Tsinghua Shenzhen International Graduate School (SIGS) and Tsinghua-Berkeley Shenzhen Institute (TBSI), Tsinghua University. He was the founding Dean of the Institute of Data and Information (iDI) at Tsinghua SIGS.

His research interests include image and multimedia content analysis, sensor networks and IoT, machine learning/AI, statistical signal processing, and applications in big data, finance, and marketing.
 
\end{IEEEbiography}

\vfill
\end{CJK}
\end{document}